\documentclass{article}
\usepackage{arxiv}

\usepackage{geometry}
\usepackage{graphicx}
\usepackage[hidelinks]{hyperref}

\usepackage{amssymb}
\usepackage{bm}
\usepackage{lineno}
\usepackage{amsmath}

\usepackage[ruled,linesnumbered]{algorithm2e}
\usepackage{cite}
\newcommand{\nonl}{\renewcommand{\nl}{\let\nl\oldnl}}
\SetKwComment{Comment}{$\triangleright$\ }{}

\usepackage{multirow}
\usepackage{float}
\usepackage{subcaption}
\usepackage{tabularx}

\usepackage{array,color}

\title{Assessment of DeepONet for reliability analysis of stochastic nonlinear dynamical systems}
\author{
  Shailesh Garg  \\
  Department of Applied Mechanics\\
  Indian Institute of Technology Delhi\\
  Hauz Khas, New Delhi 110016, India. \\
  \texttt{shaileshgarg96@gmail.com} \\
  \And
  Harshit Gupta  \\
  Department of Mechnical Engineering\\
  Indian Institute of Technology Delhi\\
  Hauz Khas, New Delhi 110016, India. \\
  \texttt{harshit.gupta.iitd23@gmail.com} \\
  \And
  Souvik Chakraborty  \\
  Department of Applied Mechanics\\
  School of Artificial Intelligence (ScAI)\\
  Indian Institute of Technology Delhi\\
  Hauz Khas, New Delhi 110016, India. \\
  \texttt{souvik@am.iitd.ac.in}}
\begin{document}
\maketitle
\begin{abstract}
Time dependent reliability analysis and uncertainty quantification of structural system subjected to stochastic forcing function is a challenging endeavour as it necessitates considerable computational time. We investigate the efficacy of recently proposed DeepONet in solving time dependent reliability analysis and uncertainty quantification of systems subjected to stochastic loading. Unlike conventional machine learning and deep learning algorithms, DeepONet learns is a operator network and learns a function to function mapping and hence, is ideally suited to propagate the uncertainty from the stochastic forcing function to the output responses. We use DeepONet to build a surrogate model for the dynamical system under consideration. Multiple case studies, involving both toy and benchmark problems, have been conducted to examine the efficacy of DeepONet in time dependent reliability analysis and uncertainty quantification of linear and nonlinear dynamical systems. Results obtained indicate that the DeepONet architecture is accurate as well as efficient. Moreover, DeepONet posses zero shot learning capabilities and hence, a trained model easily generalizes to unseen and new environment with no further training.

\end{abstract}

\section{Introduction}\label{section: Introduction}
While designing any engineering structure, performing appropriate tests to check the validity of the designs under various circumstances is of utmost importance.
For a few cases where the forces acting on the system are predictable, analysis can be performed on a fixed set of samples to verify the design of the structure.
However, in a majority of cases, the forces acting on the system are random in nature, and it is difficult for the designer to explicitly designate the design as pass or fail.
For such cases, quantifying the uncertainties in the system, assessing its influence on system responses, and performing reliability analysis \cite{rackwitz2001reliability, melchers2018structural} becomes of utmost importance.


Reliability analysis is in a way a probabilistic study \cite{lee2002comparative} of the system under consideration, and the ratio of favorable quantity to the total number of tests forms the basis of any probabilistic study.
The same is reinforced in Bayesian statistics \cite{bolstad2016introduction}, which further states that uncertainties in the final results reduce when we have a sufficiently large number of samples.
Thus to perform a robust reliability analysis in the face of uncertain forcing conditions, a large number of input force realizations are generated, and a corresponding displacement ensemble is computed.
We essentially test the system against a range of input forces to calculate its failure probability.
To perform such a comprehensive analysis, both time and computational power are required, a shortage of which can lead to inadequate and unreliable design.
Traditionally, system states for individual realizations of input force have been generated using integration schemes \cite{ruemelin1982numerical,morton2005numerical} like Runge-Kutta Method \cite{butcher1996history} and Newmark Beta method \cite{subbaraj1989survey, paz2012structural}. Techniques like Finite element modelling \cite{szabo1991finite}, MCMC simulation \cite{rubinstein2016simulation}, importance sampling \cite{engelund1993benchmark, papaioannou2016sequential} have also been used in the past.
These schemes, however accurate, carry huge computational costs, thus limiting their use to a small number of test samples.

In order to reduce the computational requirements, one may choose to decrease the dimensionality of the data using techniques like Principal Component Analysis \cite{wold1987principal} or Independent Component Analysis \cite{comon1992independent}, but however good the technique may be there is bound to be some information loss because of model order reduction \cite{schilders2008model}.
To bypass these problems, one may use machine learning \cite{mohri2018foundations,jordan2015machine} in conjunction with the traditional integration techniques.
Machine learning is a budding field of research and has found its use across several industries \cite{liu2017materials,wang2019artificial,worden2007application,wuest2016machine}.
Machine learning techniques, especially neural networks \cite{sarle1994neural,gurney2018introduction} (NN) have the ability to learn from a small training data set and then apply that same knowledge to a vast data set with very little computational cost associated with it.
Owing to this, several NN architectures have been developed with an aim to solve a set of differential equations or to learn nonlinear operators.
Popular examples include Physics Informed NN \cite{raissi2019physics}, Fourier NN \cite{silvescu1999fourier} and DeepONet \cite{lu2021learning}.

In this paper, we propose using DeepONet, a neural network architecture, to prepare a surrogate model which can then be used to generate system states for multiple test samples with reduced computational cost.
DeepONet was proposed by Lu Lu et al. in their paper \cite{lu2021learning}, and it aims at learning the nonlinear operator based on a limited training data set.
It is a data-driven architecture based on the universal approximation theorem \cite{chen1995universal} and can reliably solve ordinary differential equations and stochastic differential equations \cite{arnold1974stochastic} (SDE).
Results generated in this paper also indicate that given sufficient training data, DeepONet model demonstrates Zero-Shot Learning (ZSL) \cite{xian2018zero} capabilities for new inputs; this allows generalization of the proposed architecture.

The proposed surrogate model, while applicable for any dynamical system, is especially useful in cases where simulating high fidelity data is complex, and only a small set of the same can be generated.
Results produced show that by only simulating a few hundred samples, thousands of samples can be inferred using the surrogate model with reasonable accuracy.
It also gives the user ability to produce only relevant results i.e. if lets say displacements only at $N^{th}$ DOF are required, traditional integration schemes require the displacements and velocities to be computed at all DOFs but using the trained surrogate model, user will be able to generate only the required data thus saving expensive server space.
Additionally, for the cases where the simulation algorithm converges only for \textit{fine} time steps, the proposed surrogate model can be trained at \textit{coarser} time steps while retaining the integrity of the predictions, thus saving both computational cost and time.
While other machine learning based surrogate algorithms are available in the literature which adopt Gaussian process or Krigging \cite{atkinson2019structured,chakraborty2019graph}, NNs \cite{chakraborty2021transfer} and other such techniques \cite{chakraborty2017moment,chakraborty2017polynomial}, no work investigating the applicability of DeepONet for time-dependent reliability analysis and uncertainty quantification under forcing function uncertainty exists in the open literature, and
owing to the benefits of DeepONet listed above, it is worthwhile to investigate its applicability on the same.

The remaining paper is arranged as follows, Section \ref{section: Problem Statement} discusses the problem statement of the paper while Section \ref{section: DeepONet} introduces the DeepONet architecture being implemented. Section \ref{section: Numerical Illustration} showcases the various case studies carried out to test the proposed surrogate model. Finally, Section \ref{section: Conclusion} draws the appropriate conclusion of the paper. 
\section{Problem Statement}\label{section: Problem Statement}
Consider an $N$-DOF stochastic dynamical system with governing equation as follows:
\begin{equation}
    \mathbf M \bm{\ddot X} + \mathbf C \bm{\dot X} + \mathbf K \bm X + \bm N(\bm X, \bm{\dot X}, \bm{\theta}_N) = \bm F,
    \label{equation: stochastic dynamical system}
\end{equation}
where $\mathbf M$, $\mathbf C$ and $\mathbf K$ are the mass, damping and stiffness matrices respectively.
$\bm N(\cdot)$ is the restoring force which is a nonlinear function of system states $\bm X$ and parameter $\bm\theta_N$.
$\bm F\sim\mathbb P(\cdot)$ is the input force vector drawn from the probability distribution $\mathbb P(\cdot)$.
This uncertainty in the input force of Eq. \eqref{equation: stochastic dynamical system} lends it its stochastic nature, thus necessitating reliability analysis.

\begin{figure}[ht!]
    \centering
    \includegraphics[width = \textwidth]{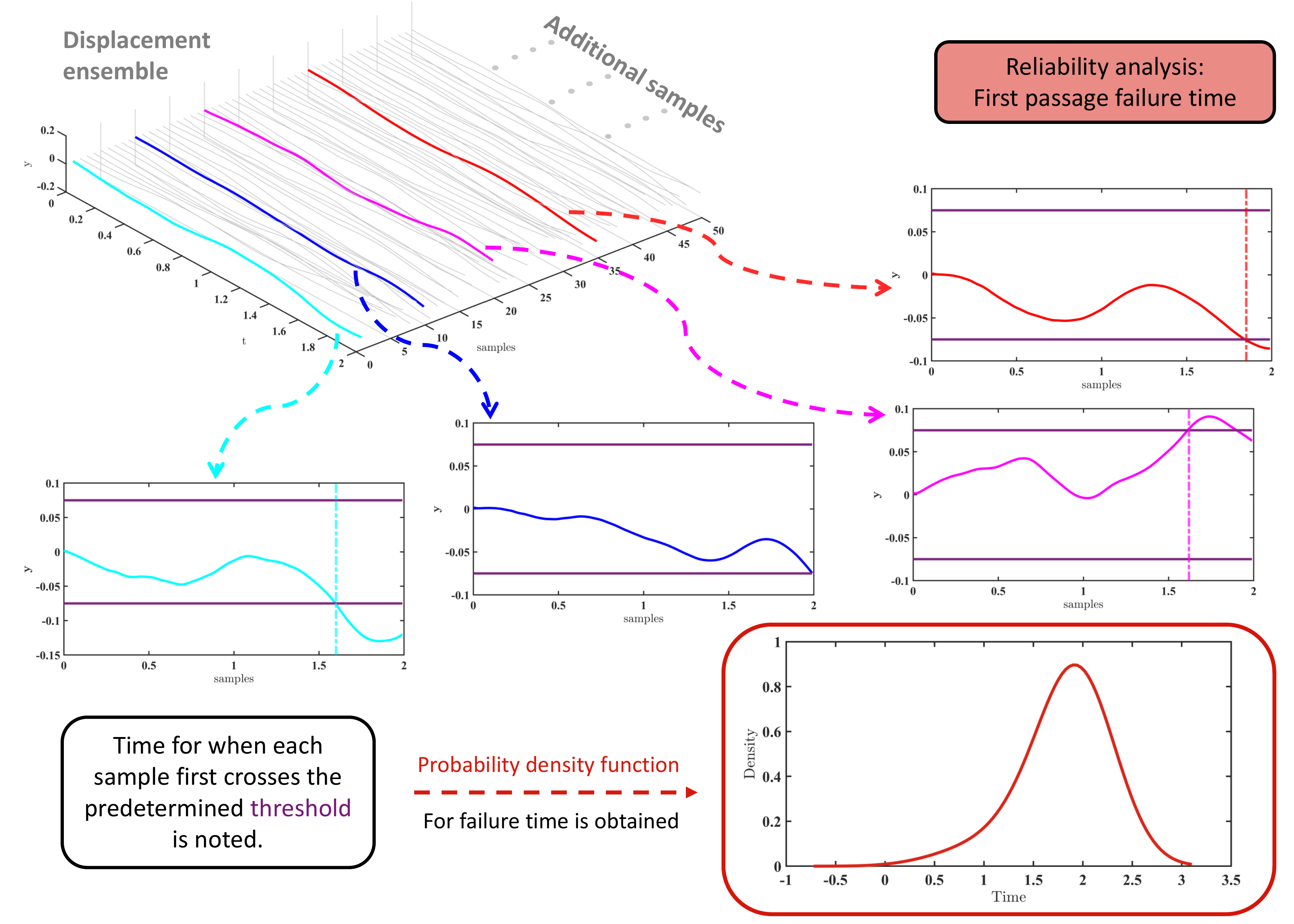}
    \caption{Pictorial representation of steps involved in reliability analysis. Specifically \textit{first passage failure time} is computed. A displacement ensemble is generated using realizations of random input force. Each sample from the resulting ensemble is tested against a predetermined threshold, and the first time the threshold is crossed is noted for each sample. If the threshold is not crossed, the final time for that sample is recorded. The probability density function for the resulting time vector is obtained, and thus the probability of failure of the structure at any given time can be obtained.}
    \label{figure: reliability analysis}
\end{figure}

\begin{figure}[ht!]
    \centering
    \includegraphics[width = \textwidth]{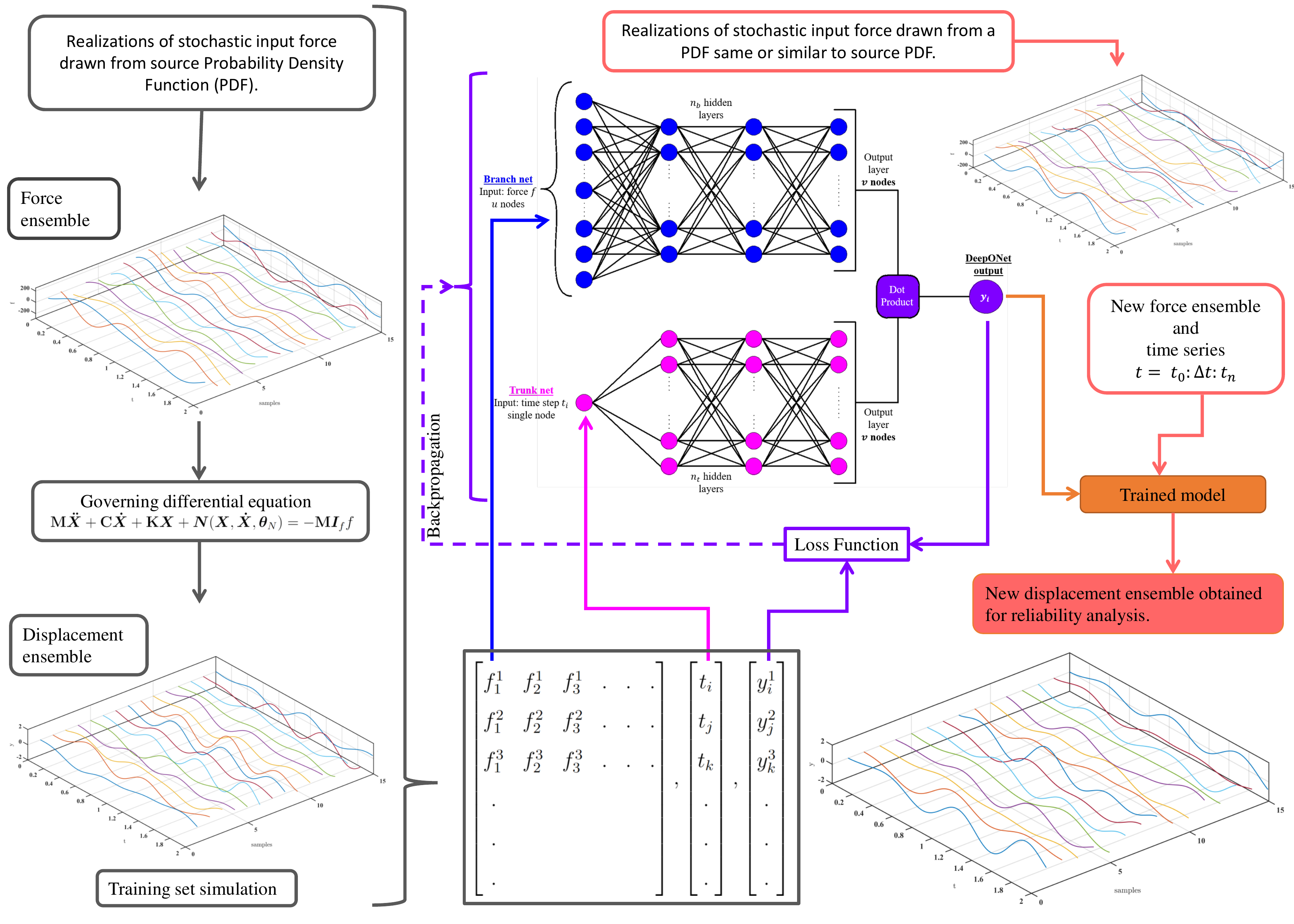}
    \caption{Schematic showcasing the process flow adopted in this paper. From a source PDF, realizations of input forces are generated, and based on the governing differential equation; displacement ensemble is obtained. The training data thus generated is rearranged as shown in Eq. \eqref{equation: deeponet dataset} and is passed through the DeepONet architecture. The trained model obtained from DeepONet is then introduced to new realizations of input force, and the corresponding displacements are generated with reduced computational cost. These displacements are then used to perform reliability analysis on the system under consideration.}
    \label{figure: detailed flow chart}
\end{figure}

Now, reliability analysis can be performed using various different techniques \cite{ditlevsen1996structural}, choice of which depends on the need of the specific application.
The first step, however is usually to draw realizations of input force from the respective probability distribution and solve Eq. \eqref{equation: stochastic dynamical system} to generate a displacement ensemble.
At which point, the methodologies diverge depending on the quantity being measured.
In this paper, we will be computing First Passage Failure Time \cite{lutes2009reliability} (FPFT), which gives a Probability Density Function (PDF) for the failure time of the structure.
Steps involved include taking realizations of displacement ensemble one by one and comparing them with the predetermined threshold.
The time step at which each sample crosses the threshold first is noted, and for samples that never cross the time step the final time is noted.
PDFs for the resulting failure time vector are computed, and further design related decisions can then be made.
The thresholds in a practical scenario are computed based on the design codes being followed and the type of structure being designed.
A schematic of the steps involved in FPFT reliability analysis is shown in Fig. \ref{figure: reliability analysis}.
As discussed earlier, the goal of this paper is to prepare a surrogate model of the Eq. \eqref{equation: stochastic dynamical system}, which can expeditiously generate results for a wide range of input forces.
This will enable us to perform reliability analysis in a short amount of time, giving a thorough picture of the expected system behavior under various loading conditions.
Steps involved in the proposed framework are given in Algorithm \ref{algorithm: Proposed framework}.
\begin{algorithm}[ht!]
\caption{Proposed Framework}\label{algorithm: Proposed framework}
Identify governing equations and PDFs of the input forces for the system under consideration.\\
Generate an ensemble of input forces and simulate training data using traditional integration schemes.\\
Train the DeepONet model using the simulated training data.\Comment*[r]{ Refer section \ref{section: DeepONet}.}
Predict displacement ensemble for new realizations of input force using the trained model. \\
Perform reliability analysis on the system under consideration using the predicted displacement ensemble.
\end{algorithm}
A detailed schematic describing the process followed is shown in Fig. \ref{figure: detailed flow chart}.
To prepare the aforementioned surrogate model, DeepONet architecture for learning nonlinear operators has been implemented.
The neural network architecture is discussed in detail in the following section.
\section{DeepONet}\label{section: DeepONet}
\begin{figure}[ht!]
    \centering
    \includegraphics[width = \textwidth]{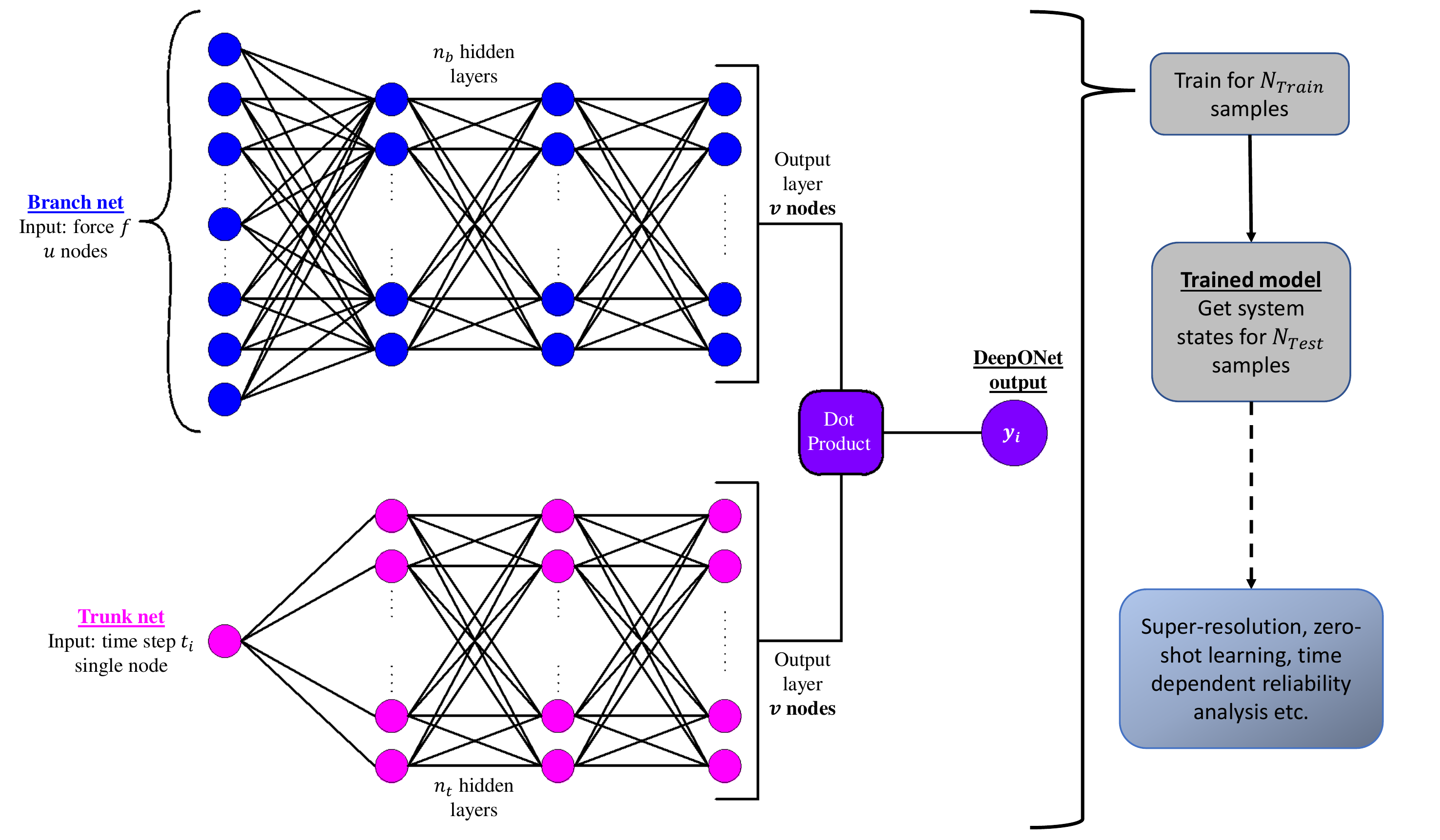}
    \caption{Flow chart showcasing the DeepONet architecture and how it has been used in the paper. DeepONet's architecture is divided into two main parts namely branch net and trunk net. The branch net takes force $f$ as input while the trunk net takes a random time step $t_i$ as input. Respective inputs are then passed through branch net and trunk net, both having \textit{same output size}. The corresponding outputs obtained are then combined and the final output in the form of displacement $y_i$ corresponding to input time step $t_i$ is obtained. Note, a normalization layer has been added in both branch and trunk net after the initial input layers.}
    \label{figure: deeponet flow chart}
\end{figure}

DeepONet is a neural network architecture developed to learn nonlinear continuous operators such as integrals, ODEs and stochastic ODEs.
It is based on the universal approximation theorem, which states that even a single layer neural network is sufficient to learn a continuous nonlinear operator.
In order to reduce the generalization error, DeepONet uses a branch and trunk architecture.
The branch network is trained to encode the input function of the nonlinear operator while the trunk network encodes the domain of the input function.
The two layers are then merged, and an output is generated.
Schematic for DeepONet architecture is shown in Fig. \ref{figure: deeponet flow chart}, mathematical representation of the same is as follows:
\begin{equation}
    G(f)(y) = \left<\alpha_b\left(b_b+\sum\limits_ix^b_iw^b_i\right)\,,\alpha_t\left(b_t+\sum\limits_ix^t_iw^t_i\right)\right>,
    \label{equation: deeponet NN dot product}
\end{equation}
where $G(f)(y)$ is the target nonlinear operator with input $f$ and output $y$. The quantities $\alpha_b\left(b_b+\sum\limits_ix^b_iw^b_i\right)$ and $\alpha_t\left(b_t+\sum\limits_ix^t_iw^t_i\right)$ are the outputs of branch and trunk network respectively.
$\alpha_b$ and $\alpha_t$ in Eq.\eqref{equation: deeponet NN dot product} represent the activation functions for the branch and trunk networks and, $b_b$ and $b_t$ are their respective bias.
$x^b$ and $x^t$ are neurons corresponding to last layer of branch and trunk network having weights $w^b$ and $w^t$ respectively.
$\left<\cdot,\cdot\right>$ represents the dot product.
Branch and trunk networks can also be configured in such a way that individually they are stacked neural networks i.e. each neuron in unstacked DeepONet will be replaced by a neural network.
In this paper, we use unstacked DeepONet architecture which takes forcing function as input for branch network and random time step as input for the trunk network.
Subsequently the output is the displacement corresponding to the input time step and the forcing function.
The training data set for the used DeepONet architecture looks as follows:
\begin{equation}
    \begin{bmatrix}
    f^1_1 & f^1_2 & f^1_3 & . & . & .\\
    f^2_1 & f^2_2 & f^2_3 & . & . & .\\
    f^3_1 & f^3_2 & f^3_3 & . & . & .\\
    .\\
    .\\
    .\\
    \end{bmatrix},
    \begin{bmatrix}
    t_i\\
    t_j\\
    t_k\\
    .\\
    .\\
    .\\
    \end{bmatrix},
    \begin{bmatrix}
    y^1_i\\
    y^2_j\\
    y^3_k\\
    .\\
    .\\
    .\\
    \end{bmatrix},
    \label{equation: deeponet dataset}
\end{equation}
where $t_i \sim \mathbb U(0,t_{end})$ and $y_i$ is the corresponding displacement.
Characters in the superscript of $f$ and $y$ represent the sample numbers while in the subscript represent the time step.
The loss function for the network is taken as the mean squared error $\epsilon$ function which is defined as follows:
\begin{equation}
    \epsilon = \frac{1}{N_p}\sum\limits_{i=1}^{N_p} (y_a-y_p)^2,
    \label{equation: MSE error deeponet}
\end{equation}
where $y_a$ is the target output of $G(f)(y)$ and $y_p$ is the output obtained from the neural network.
$N_p$ is the number of points and is equivalent to batch size in NN training.
Algorithm \ref{algorithm: deeponet algo} includes the steps involved in training the surrogate model.
\begin{algorithm}[ht!]
\caption{DeepONet Algorithm}\label{algorithm: deeponet algo}
Prepare the input data set\Comment*[r]{ Refer Eq. \eqref{equation: deeponet dataset}}
\For{\text{i = 1: no. of iterations}}{
\For{\text{j = 1: no. of training data batches}}{
Input one batch of training force into branch net.\\
Input corresponding batch of random time steps into trunk net.\\
Merge the outputs from branch and trunk net\Comment*[r]{ Refer Eq. \eqref{equation: deeponet NN dot product}}
Compute training error\Comment*[r]{ Refer Eq. \eqref{equation: MSE error deeponet}}
Update weights and biases based on back-propagation with an aim to reduce the training error.
}}
\end{algorithm}
We have implemented the DeepONet architecture using Google's Tensorflow \cite{abadi2016tensorflow} machine learning package.
\section{Numerical Illustration}\label{section: Numerical Illustration}
\begin{figure}[ht!]
    \centering
    \includegraphics[width = \textwidth]{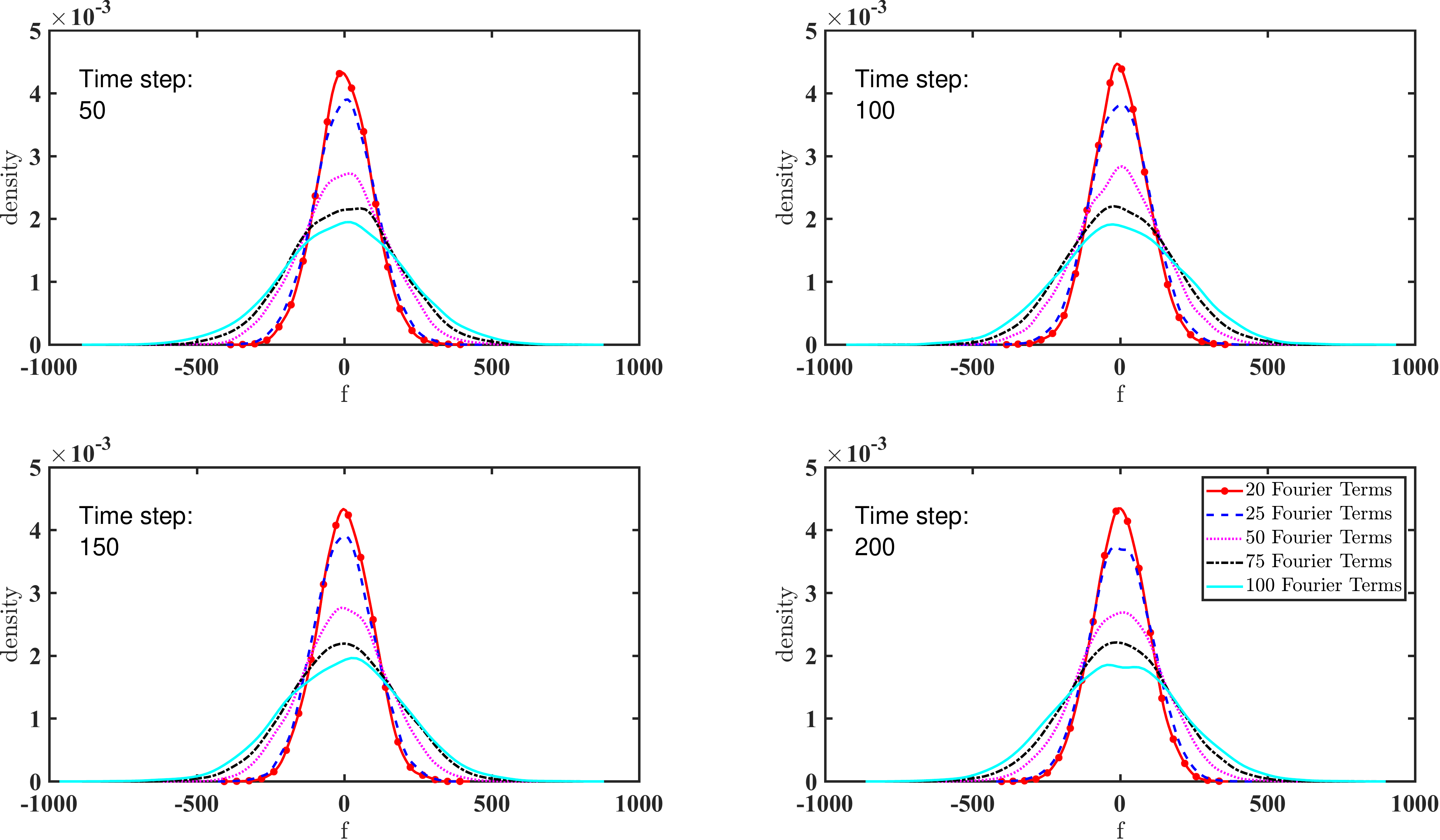}
    \caption{Probability density functions for input force computed at any particular time step. Comparisons have been drawn for cases where input force ensemble is generated using different number of FT.}
    \label{figure: force ensemble}
\end{figure}

\begin{figure}[ht!]
    \centering
    \includegraphics[width = \textwidth]{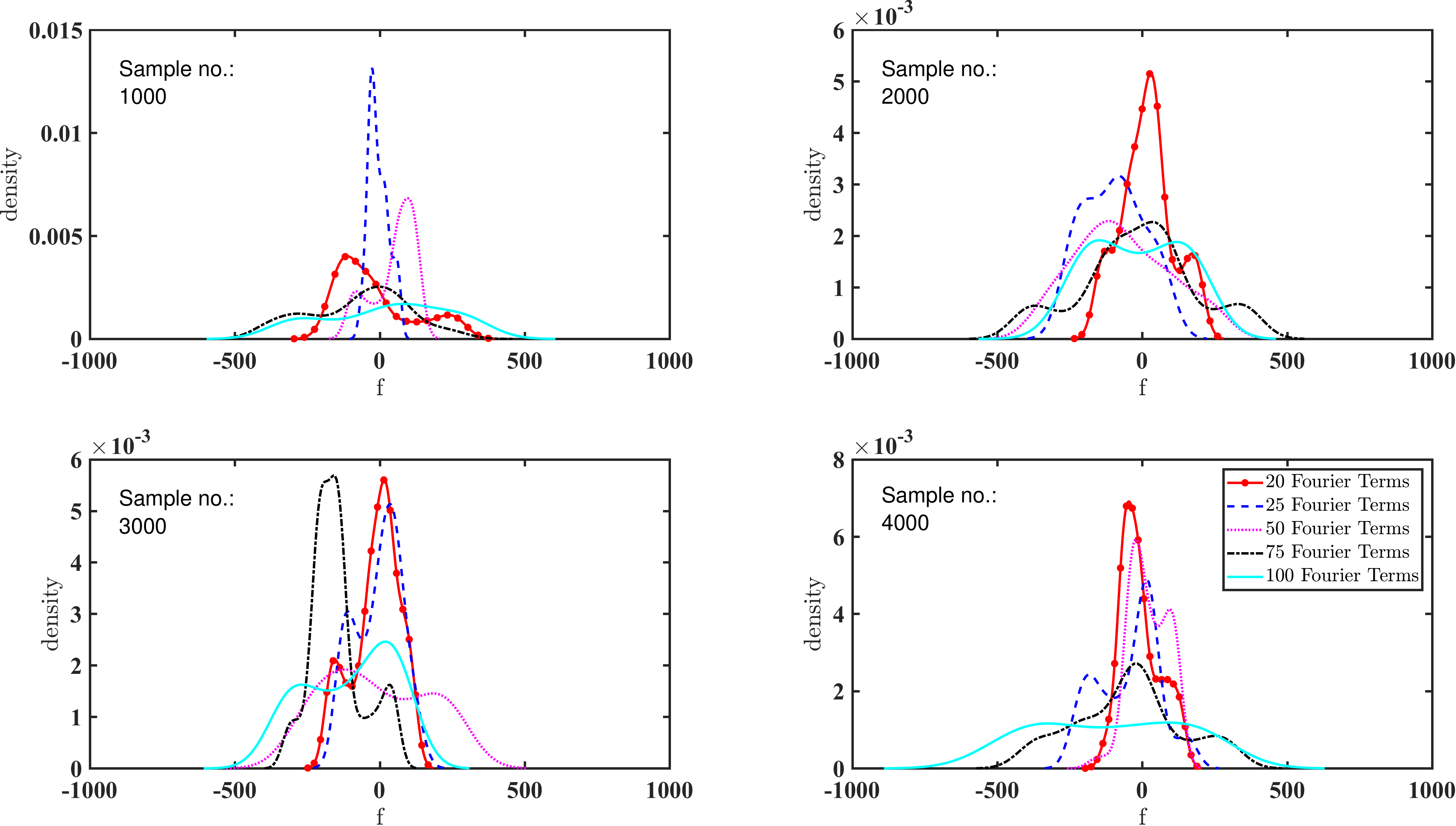}
    \caption{Probability density functions for individual samples of input force ensemble. Comparisons have been drawn for cases where input force ensemble is generated using different number of FT.}
    \label{figure: time pdf}
\end{figure}

This section covers four different case studies exploring a variety of dynamical systems.
Case-I covers a single Degree of Freedom (DOF) Bouc Wen system \cite{wen1976method,li2007nonlinear} while Case-II deals with a $5$-DOF system installed with  a duffing oscillator at first DOF.
A $76$-storey building is covered in Case-III, and the same system is modified to include a Bouc Wen oscillator at first DOF for Case-IV.
Training data for the DeepONet in the following case studies is sampled at a sampling frequency of 100$Hz$ for 2 seconds.
Input force applied in Case I(a) and II to IV is generated as follows:
\begin{equation}
    f = \sum\limits_{i = 1}^{n_s} a_{s_i} sin(f_{s_i}t)+\sum\limits_{i = 1}^{n_c} a_{c_i} cos(f_{c_i}t),
    \label{equation: force fourier}
\end{equation}
where $a_s$ and $a_c$ are the amplitudes for sine and cosine wave respectively.
$f_s$ and $f_c$ are frequencies for sine and cosine wave respectively.
$a_s$, $a_c$, $f_s$ and $f_c$ are uniform random variables where $f_s, f_c\sim \mathbb{U}(0,10)$ for all the case studies.
$a_s, a_c\sim \mathbb{U}(-50,50)$ for the first case study, $a_s, a_c\sim \mathbb{U}(-10,10)$ for the second case study and $a_s, a_c\sim \mathbb{U}(-1,1)$ for the third and fourth case study.
$n_s$ and $n_c$ in Eq. \eqref{equation: force fourier} are the number of sine and cosine terms which together $n=n_s+n_c$ are referred to as Fourier terms (FT).
Note, $n_s=n_c$ for even FT and $n_s=n_c+1$ for odd number of FT.
PDFs for the input force at a given time step, corresponding to different numbers of FT, is shown in Fig. \ref{figure: force ensemble}.
Fig. \ref{figure: time pdf} shows the PDFs of different time samples in the displacement ensemble. 
Forcing function for Case I(b) is drawn from a Gaussian Process having squared exponential as its kernel function.
Parameter details for the kernel and mean function are discussed in detail in the case study itself.

DeepONet architecture selected for the following case studies is as follows:
\begin{enumerate}
    \item \textbf{Branch net:} 100 $\rightarrow$ Batch Normalization $\rightarrow$ 40 $\rightarrow$ 40.
    \item \textbf{Trunk net:} 1 $\rightarrow$ Batch Normalization $\rightarrow$ 40 $\rightarrow$ 40.
    \item \textbf{Output:} Dot product $\rightarrow$ 1.
\end{enumerate}
All layers except the final output layer with single node have ReLu activation function \cite{agarap2018deep}. Glorot normal function \cite{glorot2010understanding} is used for initializing the weights. Unlike conventional ML algorithms, DeepONet generates function to function mapping (instead of variables to variables);
this is why input for branch net is a realization of input force, which has been discretized into 100 points, thus necessitating 100 nodes in the input layer of branch net.
The input size will change if forcing function is provided at finer or coarser intervals.
The choice of number of hidden layer nodes is subjective and will change according to the needs of the application.
For the following case studies however, nodes in hidden layers have been kept same.
\subsection{Case-I: SDOF Bouc Wen System}
For the first case study, the applicability of the surrogate model is tested for an SDOF Bouc Wen system.
Governing equations for the system under consideration are as follows:
\begin{equation}
\begin{matrix}
m\ddot y  + c\dot y + ky + (1-k_r)Q_yz = f\\
\dot z = \frac{1}{D_y}(\alpha\dot y - \gamma z|\dot y||z|^{n-1} - \beta\dot y|z|^n)
\end{matrix}
\label{equation: sdof bw}
\end{equation}
where $m$, $c$ and $k$ are the mass, stiffness and damping matrices. $Q_y$, $k_r$, $\alpha$, $\beta$, $\gamma$, $D_y$ and $\eta$ are the parameters of the Bouc Wen oscillator.
Readers interested to read more about Bouc Wen oscillator may refer \cite{wen1976method,li2007nonlinear}.
System parameters are selected from \cite{shirali2009principal} and are given in Table \ref{table: sdof BW system parameters}.
\begin{table}[ht!]
	\centering{
		\begin{tabular}{|c|c|c|c|}
			\hline
			 Mass (Kg) & Stiffness (N/m) & Damping (Ns/m) & Nonlinear Parameters \\ \hline
			 $m = 6800$ & $k = 232000$ & $c = 3750$ & $\begin{matrix}Q_y = 0.05mg, k_r = \frac{1}{6}, \alpha = 1,\\ \beta = 0.5, \gamma = 0.5, D_y = 0.0013, \eta =2\end{matrix}$ \\\hline
	\end{tabular}}
	\caption{System parameters for Case-I: SDOF Bouc Wen Oscillator.}
	\label{table: sdof BW system parameters}
\end{table}
\subsubsection{Case-I (a)}
Input forces for part \textit{a} of Case-I have been generated using Eq. \eqref{equation: force fourier} with 20 FT.
Testing and training data have been simulated based on Eq. \eqref{equation: sdof bw} and initial conditions have been taken as $y_0 = 0.01,\,\dot y_0 = 0.05$.
Comprehensive testing has been carried out to test the surrogate model against a variety of setups and develop an intuition as to what works and why.

The Mean square error (MSE) values in Table \ref{table: MSE Loss for Case-I varying samples} showcase the performance of surrogate model when trained for varying number of training samples.
Initial samples vary from 25 to 150 with time steps per sample or points per sample (PPS) fixed at 100; implying that in case of 150 initial samples, displacements were simulated for 150 different input forces, and for each displacement 100 uniformly distributed time steps were serially selected for training.
Once trained, displacement ensemble was generated using 10000 different input forces, and the DeepONet predictions were compared against the ground truth to get the MSE.
It can be observed that as the number of initial samples increases, the MSE value decreases which is a direct effect of increased training data.
In the same table, the performance of the trained model is also shown for cases where input forces are generated using different number of FT.
As we increase the number of FT from 20 to 100 in prediction stage, the MSE values increase, but they are still negligible for the case with only 150 initial samples. This showcases ZSL capabilities of the surrogate model as it is able to produce good results for input distribution dissimilar to the one used for training.
\begin{table}[ht!]
\centering
\begin{tabular}{|c|c|c|c|c|c|c|}
\hline
\multicolumn{2}{|c|}{\multirow{2}{*}{M.S.E.}} & \multicolumn{5}{c|}{FT} \\ \cline{3-7} 
\multicolumn{2}{|c|}{} & 20 & 25 & 50 & 75 & 100 \\ \hline
\multicolumn{1}{|c|}{\multirow{5}{*}{Samples}} & 25 & 5.54E-07 & 7.28E-07 & 2.25E-06 & 4.10E-06 & 6.09E-06\\ \cline{2-7}
\multicolumn{1}{|c|}{} & 50 & 7.33E-08 & 9.08E-08 & 2.58E-07 & 5.90E-07 & 1.03E-06 \\ \cline{2-7}
\multicolumn{1}{|c|}{} & 75 & 7.86E-08 & 1.18E-07 & 5.02E-07 & 1.00E-06 & 1.61E-06 \\ \cline{2-7}
\multicolumn{1}{|c|}{} & 100 & 7.11E-08 & 8.99E-08 & 2.67E-07 & 5.51E-07 & 9.01E-07 \\ \cline{2-7}
\multicolumn{1}{|c|}{} & 150 & 9.50E-09 & 1.08E-08 & 2.77E-08 & 7.04E-08 & 1.48E-07 \\ \hline
\end{tabular}
\caption{MSE values comparing performance of surrogate model when trained for different number of initial samples, for Case-I (a). PPS are fixed at 100. Comparison is also drawn for case when tested for input corresponding to different number of FT. Note, that the training is done for 20 FT only.}
\label{table: MSE Loss for Case-I varying samples}
\end{table}

\begin{figure}[ht!]
    \centering
    \includegraphics[width = \textwidth]{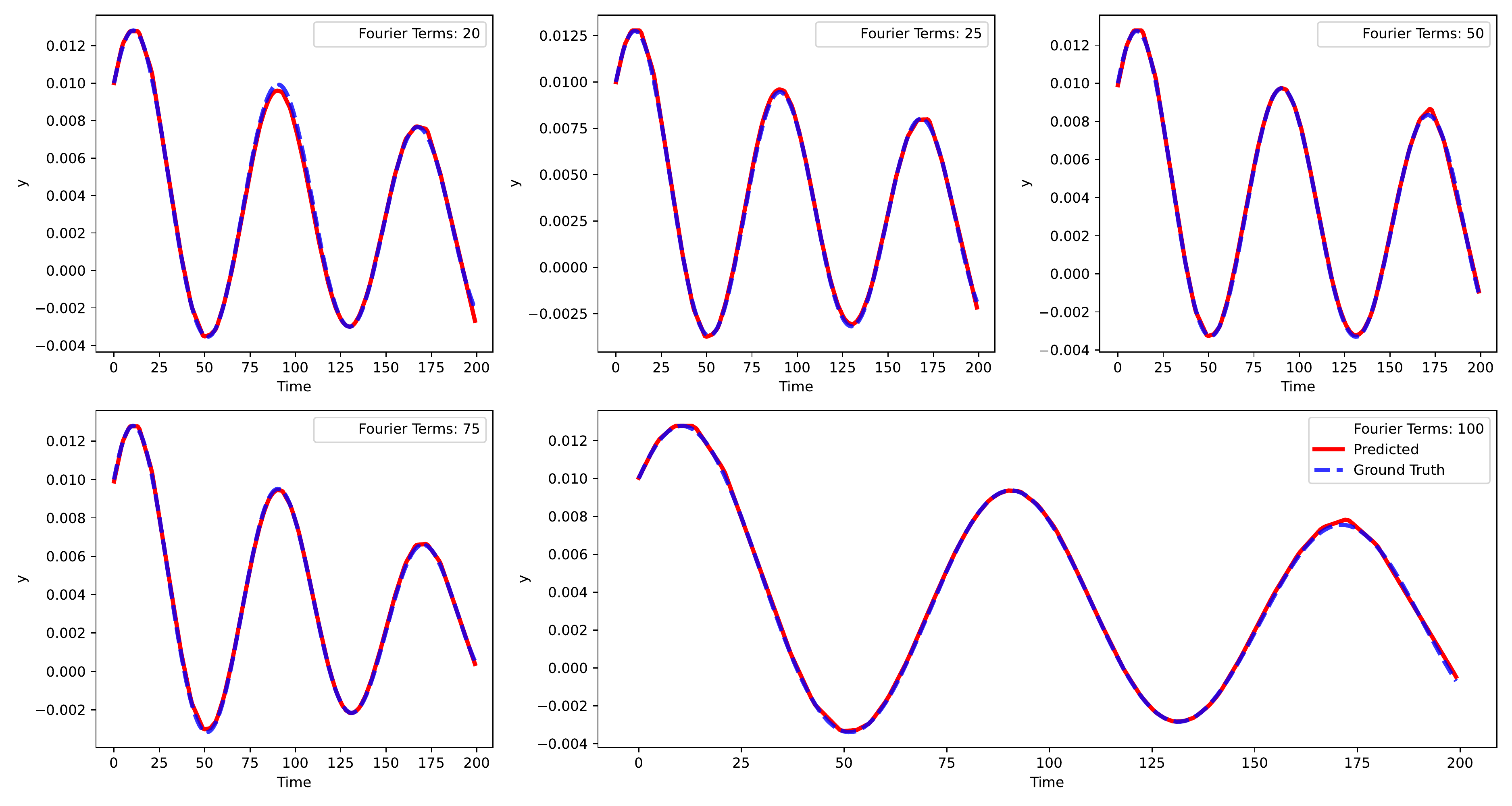}
    \caption{Predicted displacement for $1000^{th}$ sample compared against the ground truth, for Case-I (a). 150 initial samples are taken with 100PPS. Training is done for input forces corresponding to 20 FT.}  
    \label{figure: varying samples SDOF BW sample 1000}
\end{figure}

Fig. \ref{figure: varying samples SDOF BW sample 1000} shows the results produced for sample number 1000 and Fig. \ref{figure: mean var SDOF BW, 150 training samples} shows the ensemble mean, and variance of 10000 samples plotted against the ground truth when the model is trained using 150 initial samples having 100 PPS.
The plots further validate the trend shown by the MSE values as the predicted values closely follow the ground truth.
\begin{figure}[ht!]
    \centering
    \includegraphics[width = \textwidth]{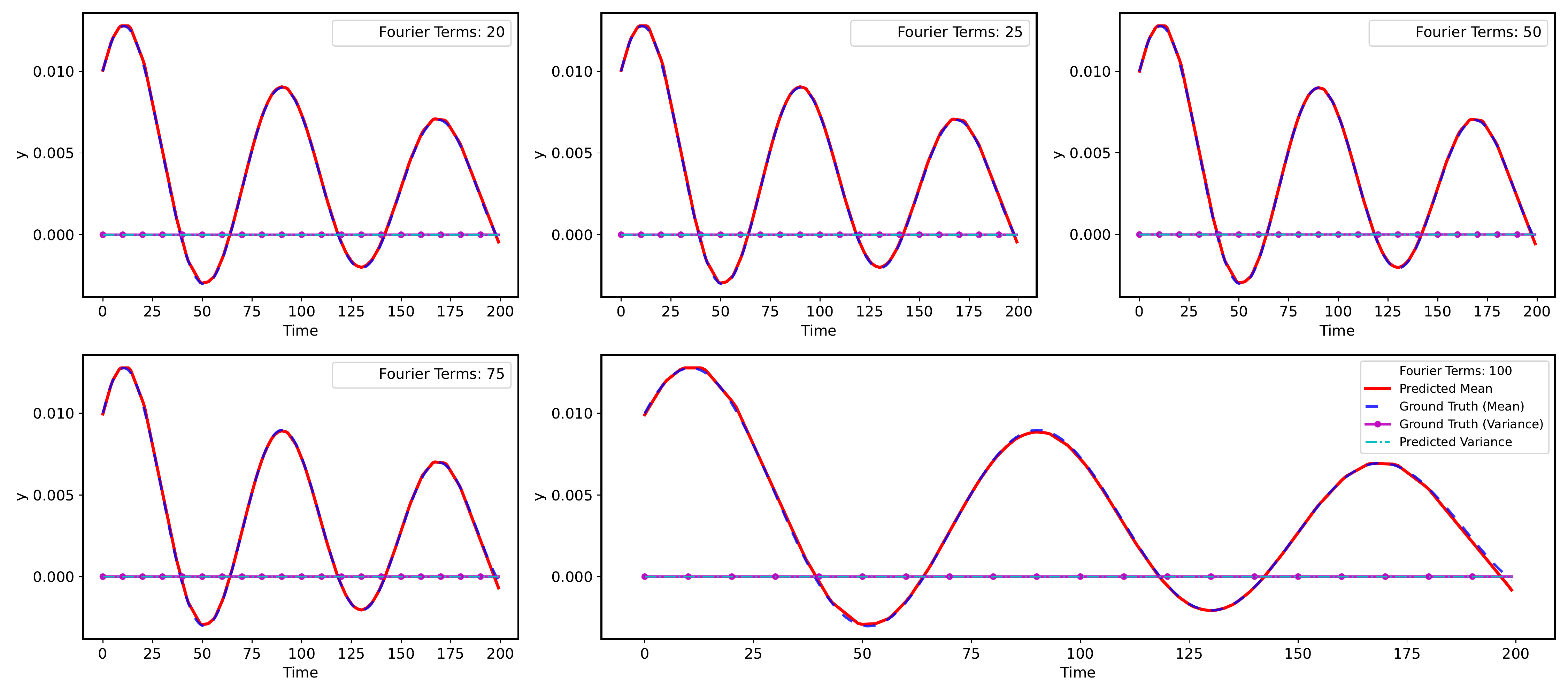}
    \caption{Predicted mean and variance of 10000 samples compared against the ground truth, for Case-I (a). 150 initial samples are taken with 100PPS. Training has been done for input forces corresponding to 20 FT.}
    \label{figure: mean var SDOF BW, 150 training samples}
\end{figure}

Now that the variation in performance is tested for changing initial samples, another avenue of testing will be exploring the effects of changing PPS. 
Table \ref{table: MSE SDOF BW varying PPS} shows the MSE values for varying PPS, keeping the number of initial samples fixed at 150.
It can be observed that as the number of PPS increases, the MSE value decreases.
This is simply due to the fact that while training, if PPS are more, surrogate model will get a better representation of the output function, even when initial samples are kept same.
Best performance among the tested PPS is observed for PPS equal to 100.
It should be noted that in Tables \ref{table: MSE Loss for Case-I varying samples} and \ref{table: MSE SDOF BW varying PPS} initial samples are as low as 50 with 100 PPS and PPS as low as 25 with initial samples as 150 show negligible MSE values.
The choice of 150 initial samples with 100 PPS however gave the best results for ZSL.
If the goal is only to perform predictions within the training distribution, fewer initial samples and/or PPS may also give good results.
\begin{table}[ht!]
\centering
\begin{tabular}{|c|c|c|c|c|c|c|}
\hline
\multicolumn{2}{|c|}{\multirow{2}{*}{M.S.E.}} & \multicolumn{5}{c|}{FT} \\ \cline{3-7} 
\multicolumn{2}{|c|}{} & 20 & 25 & 50 & 75 & 100 \\ \hline
\multicolumn{1}{|c|}{\multirow{5}{*}{PPS}} & 1 & 6.11E-04 & 7.78E-04 & 1.69E-03 & 2.60E-03 & 3.47E-03\\ \cline{2-7}
\multicolumn{1}{|c|}{} & 25 & 9.16E-08 & 1.09E-07 & 2.65E-07 & 4.67E-07 & 8.72E-07 \\ \cline{2-7}
\multicolumn{1}{|c|}{} & 50 & 2.46E-08 & 3.58E-08 & 1.65E-07 & 4.28E-07 & 8.18E-07 \\ \cline{2-7}
\multicolumn{1}{|c|}{} & 75 & 1.53E-07 & 2.08E-07 & 6.86E-07 & 1.47E-06 & 2.52E-06 \\ \cline{2-7}
\multicolumn{1}{|c|}{} & 100 & 1.66E-08 & 2.32E-08 & 1.09E-07 & 2.52E-07 & 4.92E-07 \\ \hline
\end{tabular}
\caption{MSE values comparing performance of surrogate model when trained for different number of PPS, for Case-I (a). Initial samples are fixed at 150. Comparison is also drawn for case when tested for input corresponding to different number of FT. Note, that the training is done for 20 FT only.}
\label{table: MSE SDOF BW varying PPS}
\end{table}
\subsubsection{Case-I (b)}
For part \textit{b}, input forces are taken as realizations of Gaussian process, which is defined as follows:
\begin{equation}
    f\sim\mathcal{GP}(\mu(\bm\theta_\mu),\kappa(\bm\theta_\kappa)),
\end{equation}
where $\mathcal GP(\cdot)$ is the Gaussian process with $\mu(\cdot)$ as its mean function and $\kappa(\cdot)$ as its kernel function.
$\bm\theta_\mu$ and $\bm\theta_\kappa$ are the parameters for the mean and kernel function respectively.
In this instance, $f$ is taken as a zero mean Gaussian process with squared exponential function as its kernel,
\begin{equation}
    \kappa(t,t') = \sigma^2exp\left({-\frac{(t-t')^2}{2l^2}}\right),\hspace{2em}\bm\theta_\kappa = [\sigma, l]
\end{equation}
where $\sigma$ is the variance parameter and $l$ is the length scale parameter.
Initial conditions for data simulation are taken as $y_0 = 0.005$ and $\dot y_0 = 0.001$.
For training the surrogate model variance parameter is taken as $\sigma = 50$ and the length scale parameter is taken as $l = 0.10$.
1000 initial samples are taken for training the surrogate model with 50 PPS.

\begin{figure}[ht!]
    \centering
    \includegraphics[width = \textwidth]{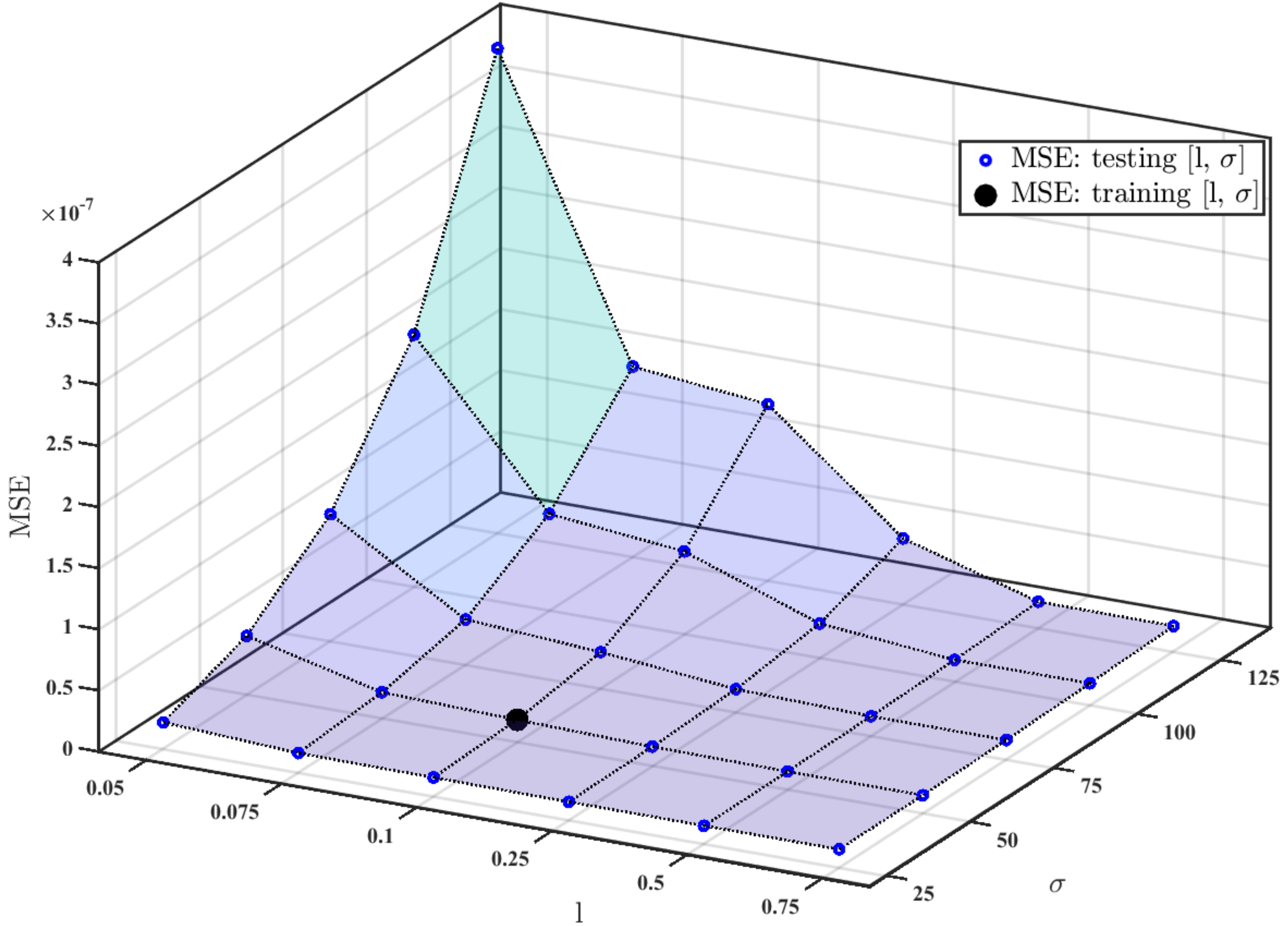}
    \caption{MSE values comparing performance of surrogate model when tested for different combinations of $\sigma$ and $l$, for Case-I (b). Initial samples are fixed at 1000 with 50 PPS. Note, that the training is done for $\sigma = 50$ and $l = 0.10$}
    \label{figure: MSE GP SDOF BW}
\end{figure}

The surrogate model is tested for various
combinations of $\sigma$ and $l$ and Fig. \ref{figure: MSE GP SDOF BW} shows the MSE of 10000 samples for each combination.
It can be observed that while the MSE values are higher for length scale below 0.10, they are of the order $10^{-7}$.
Similarly, for values of $\sigma$ more than 50, MSE values increase but stay within reasonable range.
This shows the ZSL capabilities of the proposed framework.
Equally good results were observed for 750 initial samples but are not included in the interest of avoiding recapitulation of similar results.

\begin{figure}[ht!]
    \centering
    \includegraphics[width = \textwidth]{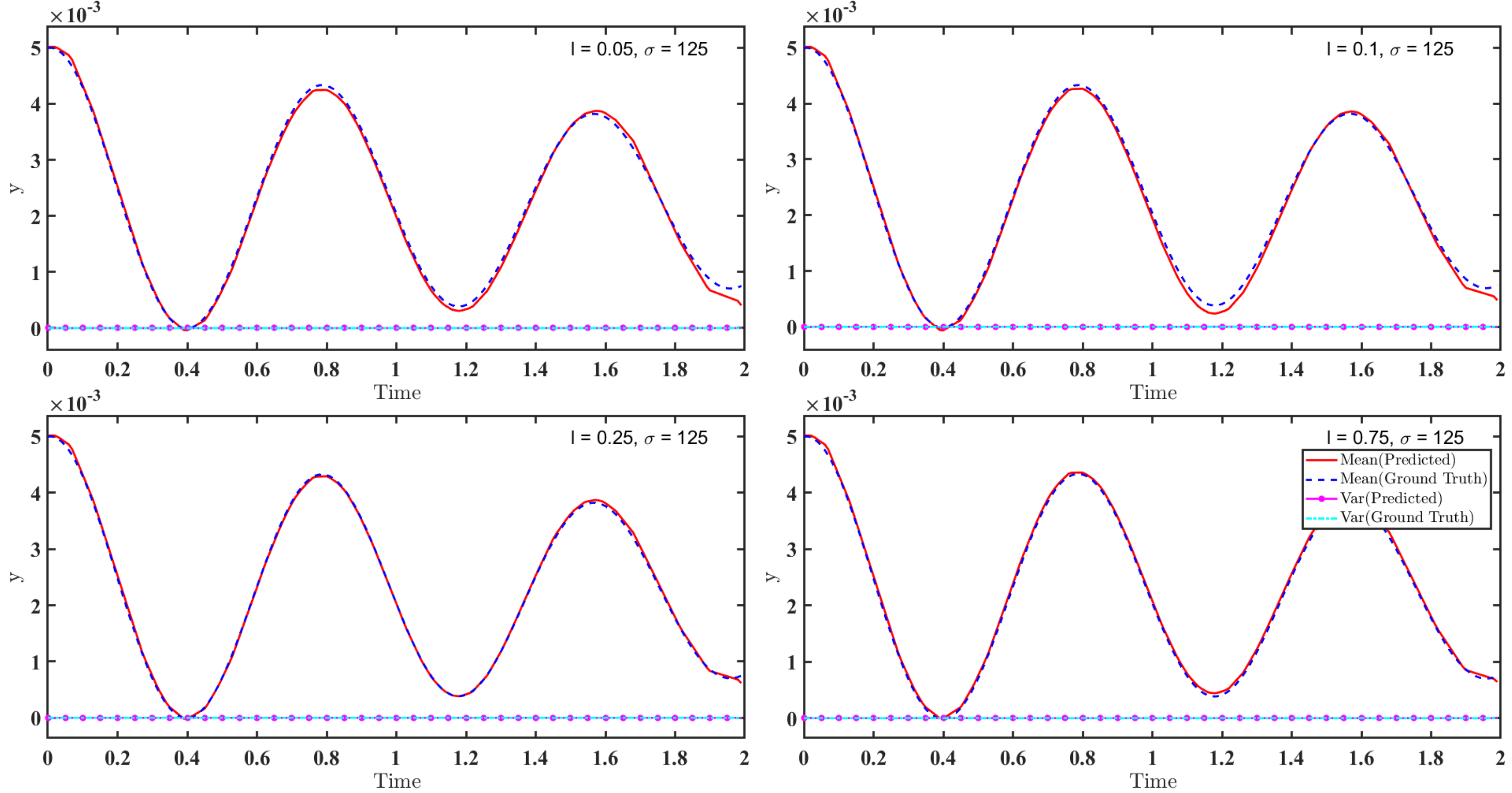}
    \caption{Predicted mean and variance of 10000 samples compared against the ground truth, for Case-I (b). 1000 initial samples are taken with 50PPS. Training is done for input forces corresponding to $\sigma = 50$ and $l = 0.10$.}
    \label{figure: mean var GP SDOF BW}
\end{figure}

Fig. \ref{figure: mean var GP SDOF BW} shows the predicted mean and variance for 10000 test samples compared against the ground truth.
As a detailed visual confirmation for the MSE values plotted earlier, comparisons have been drawn for four different combinations of $\sigma$ and $l$, and as expected, predictions closely follow the ground truth.

In this case study, the behavior of the dynamical system was dominated by the initial conditions selected and the parameters of the nonlinear oscillator, but in the subsequent case studies, we will see the performance of the surrogate model for nonlinear multi DOF systems with non-zero initial conditions; whose behavior is rather significantly affected by the input force.
Also, since the maximum displacement is observed at initial time steps only, reliability analysis is skipped for this system.
\subsection{Case-II: $5$-DOF Nonlinear System}
\begin{figure}[ht!]
    \centering
    \includegraphics[width = \textwidth]{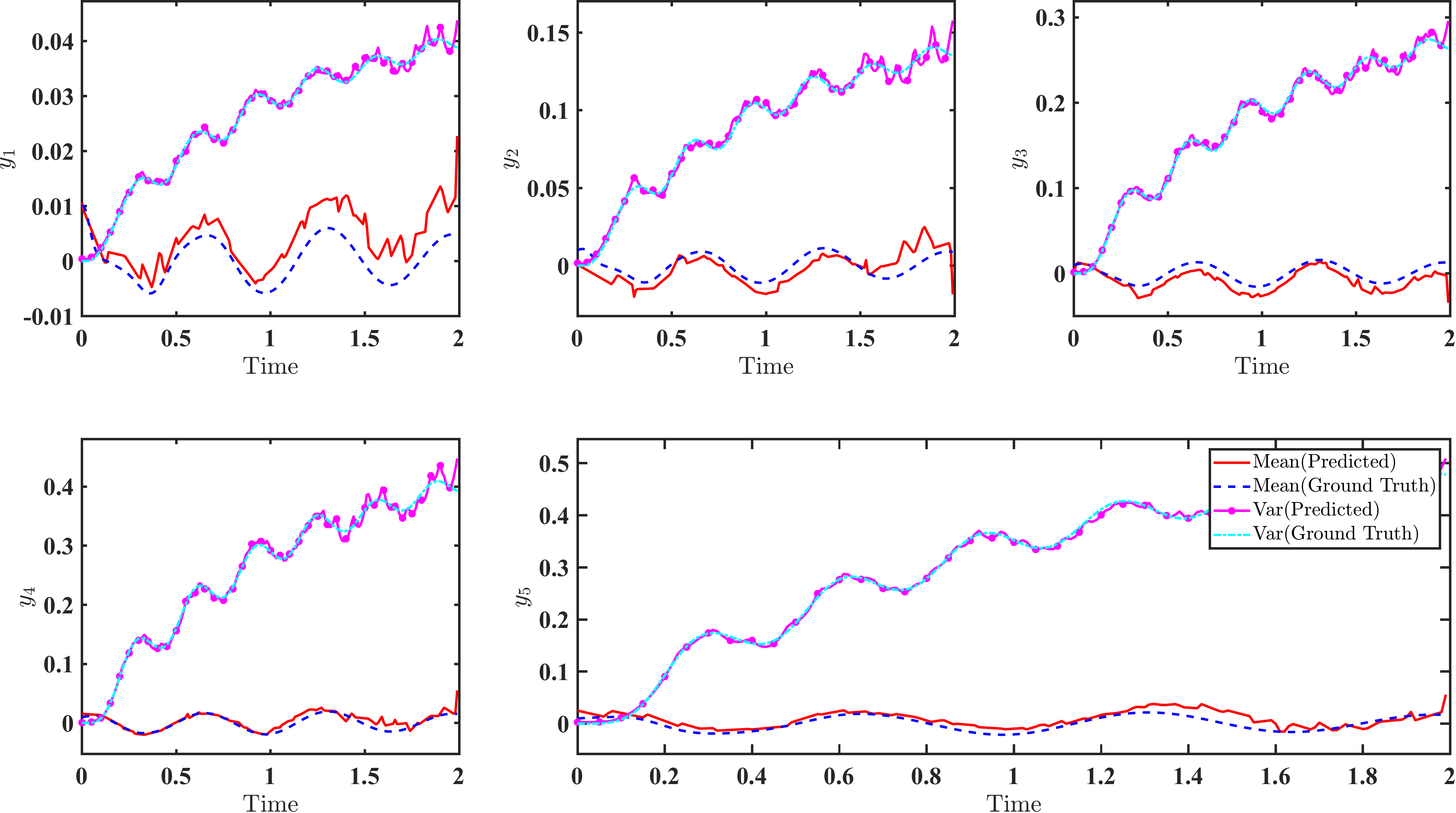}
    \caption{Predicted mean and variance of 10000 samples compared against the ground truth, for Case-II. 500 initial samples are taken with 25PPS. Training and testing has been done for input forces corresponding to 20 FT.}
    \label{figure: 5DOF_DO_mean_var_FT_20_samples_500}
\end{figure}

\begin{figure}[ht!]
    \centering
    \includegraphics[width = \textwidth]{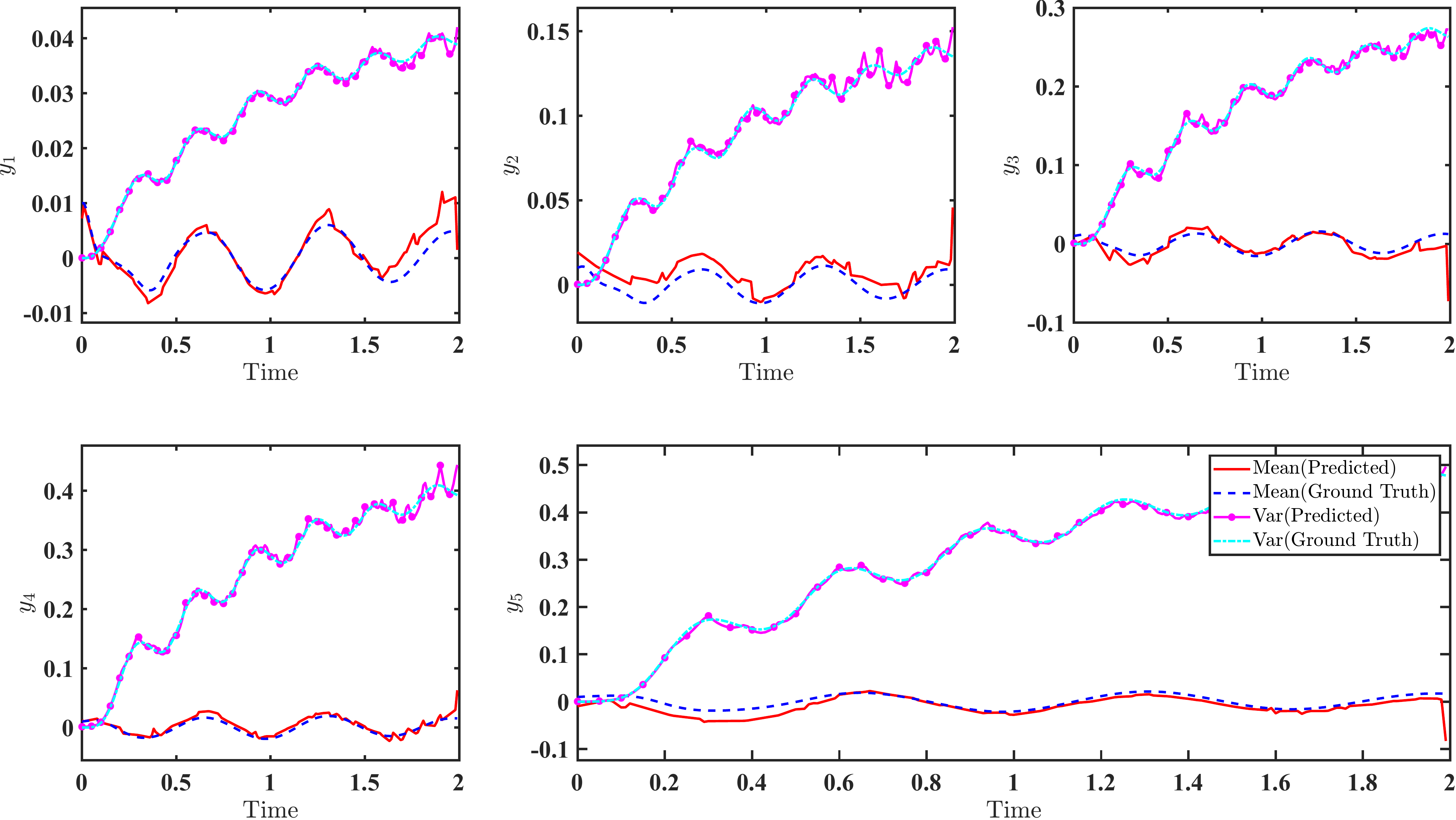}
    \caption{Predicted mean and variance of 10000 samples compared against the ground truth, for Case-II. 1500 initial samples are taken with 25PPS. Training and testing has been done for input forces corresponding to 20 FT.}
    \label{figure: 5DOF_DO_mean_var_FT_20_samples_1500}
\end{figure}

This case study deals with a base excited $5$-DOF system with duffing oscillator installed at first DOF, governing equations of which are as follows:
\begin{equation}
    \begin{matrix}
      m_1\ddot x_1+c_1\dot x_1+c_2(\dot x_1-\dot x_2)+k_1x_1+k_2(x_1-x_2)+\alpha_{do}x_1^3= -m_1f\\
      m_2\ddot x_2+c_2(\dot x_2-x_1)+c_3(\dot x_2-x_3)+k_2(\dot x_2-x_1)+k_3(\dot x_2-x_3) = -m_2f\\
      m_3\ddot x_3+c_3(\dot x_3-x_2)+c_4(\dot x_3-x_4)+k_3(\dot x_3-x_2)+k_4(\dot x_3-x_4) = -m_3f\\
      m_4\ddot x_4+c_4(\dot x_4-x_3)+c_5(\dot x_4-x_5)+k_4(\dot x_4-x_3)+k_5(\dot x_4-x_5) = -m_4f\\
      m_5\ddot x_5+c_5(\dot x_5-x_4)+k_5(\dot x_5-x_4) = -m_5f,\\
      x_i(0) = 0.01,\,\,\dot x_i(0) = 0.05,\,\,i = [1, 2, 3, 4, 5]  
    \end{matrix}
    \label{equation: duffing oscillator 5dof}
\end{equation}
where $m_i$, $c_i$ and $k_i$ are the mass, damping and stiffness parameters respectively and $\alpha_{do}$ is the constant for duffing oscillator.
$f$ in this case study represents the ground acceleration, and randomness of input force is derived from the randomness of ground acceleration.
Henceforth in Case-II to IV, ground acceleration and input force have been used interchangeably because in all three case studies, base excited systems have been considered.
System parameters for the 5DOF system shown in Eq. \eqref{equation: duffing oscillator 5dof} are given in Table \ref{table: 5dof system parameters}.
\begin{table}[ht!]
	\centering{
		\begin{tabular}{|c|c|c|c|}
			\hline
			 Mass (Kg) & Stiffness (N/m) & Damping (Ns/m) & nonlinear Parameters \\ \hline
			 $\begin{matrix}m_1 = 10, m_2 = 10\\m_3 = 9, m_4 = 9\\m_5 = 7.5\end{matrix}$ & $\begin{matrix}k_1 = 10000,k_2 = 10000\\k_3 = 9000,k_4 = 9000\\k_5 = 7500\end{matrix}$ & $\begin{matrix}c_1 = 100,c_2 = 100\\c_3 = 90,c_4 = 90\\c_5 = 75\end{matrix}$ & $\alpha = 100$ \\\hline
	\end{tabular}}
	\caption{System parameters for Case-II: $5$-DOF Duffing Oscillator.}
	\label{table: 5dof system parameters}
\end{table}
DeepONet architecture similar to previous case has been adopted but for multi DOF systems, during training, $f$ is mapped to displacements in parallel i.e. for this case study 5 trained models will be available at the end of analysis.
PPS are taken equal to 25 and performance of surrogate model is tested for 500 and 1500 initial samples.
Figs. \ref{figure: 5DOF_DO_mean_var_FT_20_samples_500} and \ref{figure: 5DOF_DO_mean_var_FT_20_samples_1500} show the performance of surrogate model for different number of initial samples (500 and 1500 respectively) by comparing the mean and variance of the predicted displacements against the ground truth.
10000 test samples were considered while plotting each result.
Predicted values closely follow the ground truth even for 500 samples and as expected, the results improve as the number of training samples increases.

\begin{figure}[ht!]
    \centering
    \includegraphics[width = \textwidth]{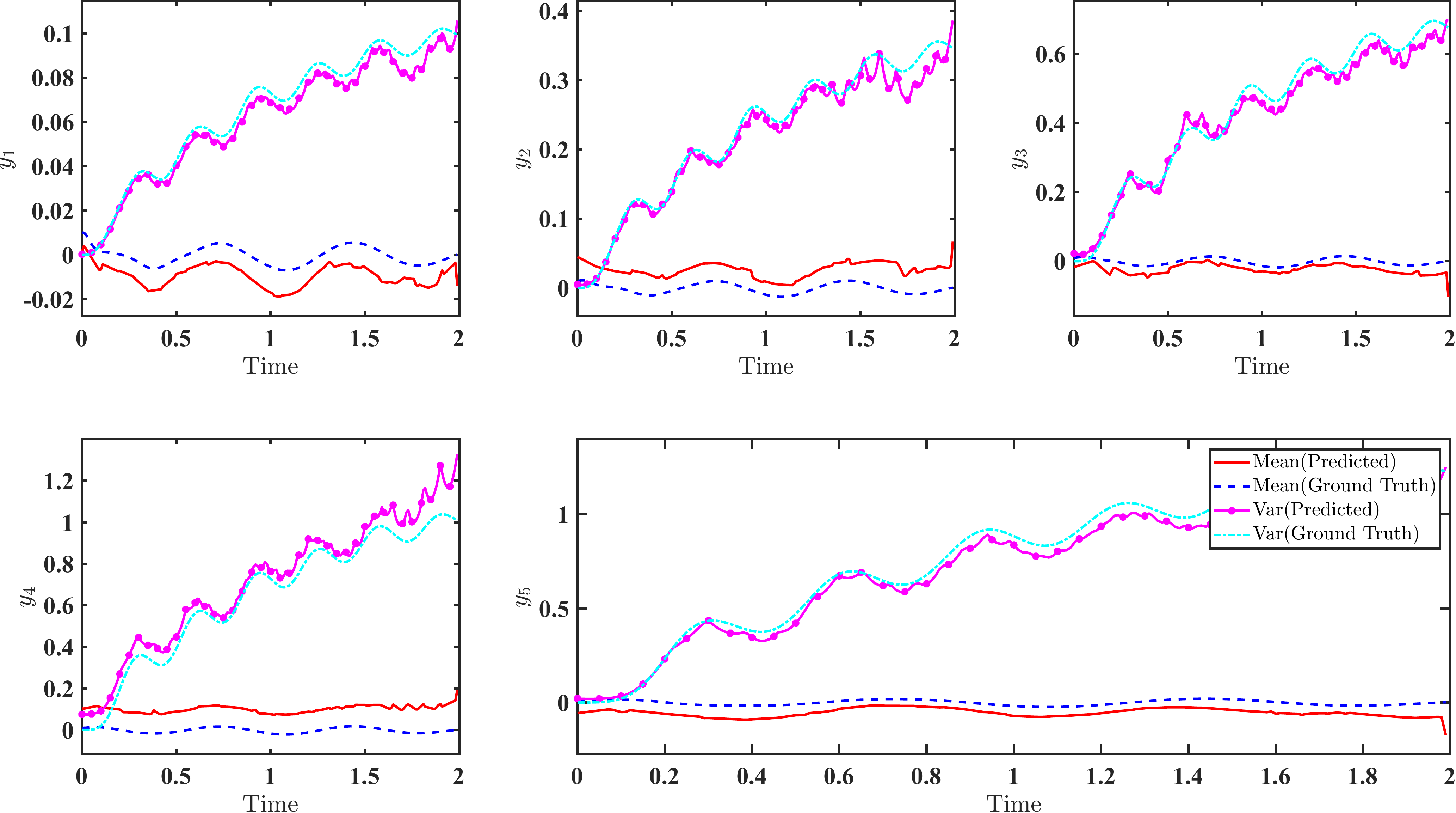}
    \caption{Predicted mean and variance of 10000 samples compared against the ground truth, for Case-II. 1500 initial samples are taken with 25PPS. Training is done for input forces corresponding to 20 FT and testing has been done for input forces corresponding to 50 FT.}
    \label{figure: 5DOF_DO_mean_var_FT_50_samples_1500}
\end{figure}

\begin{figure}[ht!]
    \centering
    \includegraphics[width = \textwidth]{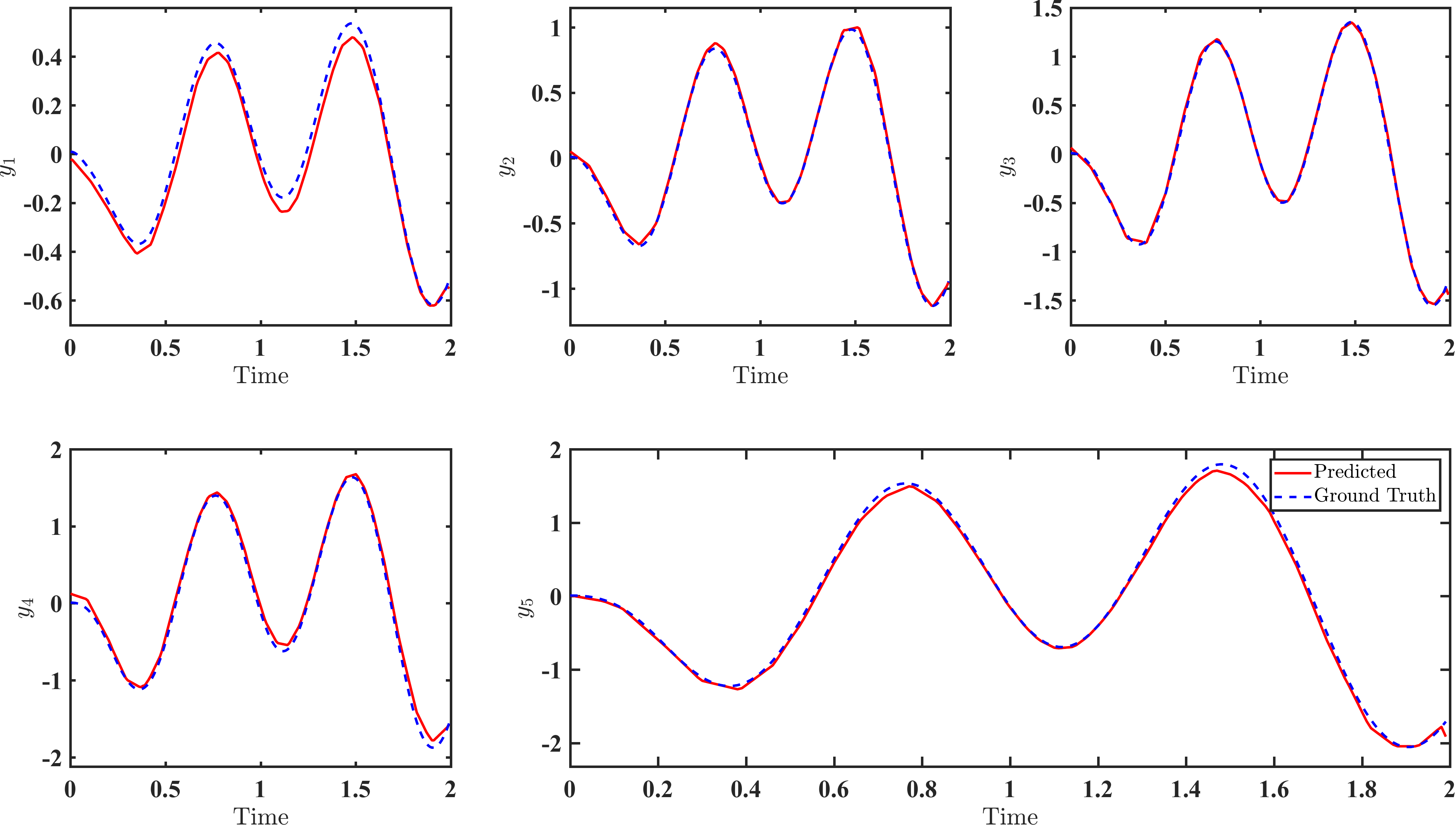}
    \caption{Predicted displacement for $5000^{th}$ sample compared against the ground truth, for Case-II. 1500 initial samples are taken with 25PPS. Training is done for input forces corresponding to 20 FT and testing has been done for input forces corresponding to 50 FT.}
    \label{figure: 5DOF_DO_Sno_5000_FT_50_samples_1500}
\end{figure}

To show the ZSL capabilities of the surrogate model, testing results for input force corresponding to 50FT are plotted in Fig. \ref{figure: 5DOF_DO_mean_var_FT_50_samples_1500}. The mean and variance of 10000 samples are shown, and good approximations of ground truth are obtained. It should be noted that the training here was done for input forces corresponding to 20 FT, with 1500 initial samples.
To further support the mean and variance results, Fig. \ref{figure: 5DOF_DO_Sno_5000_FT_50_samples_1500} shows the $5000^{th}$ sample for case when the model is trained for 20 FT and predictions are made for 50 FT.
From Fig. \ref{figure: 5DOF_DO_Sno_5000_FT_50_samples_1500}, it can be inferred that the minor gap observed in predicted mean and ground truth in the Fig. \ref{figure: 5DOF_DO_mean_var_FT_50_samples_1500} is trivial as compared to the actual magnitude of the displacement and thus has a negligible effect on the final predictions.

\begin{figure}[ht!]
    \centering
    \includegraphics[width = \textwidth]{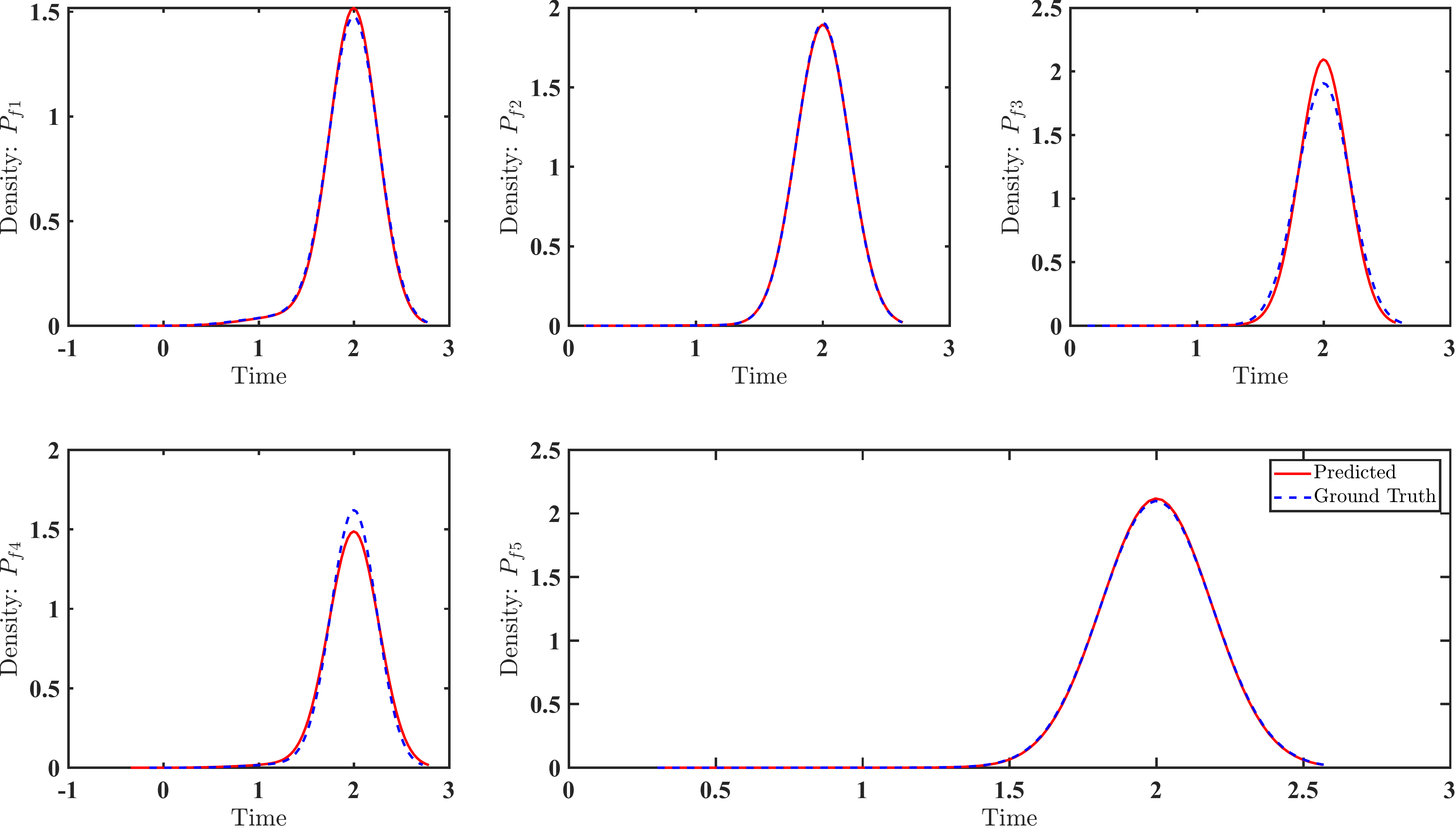}
    \caption{PDFs of FPFT obtained for Case-II. Training and testing are done for input forces corresponding to 20FT.}
    \label{figure: FPFT 5DOF DO}
\end{figure}

Now that the predictions are available for a large number of testing samples, FPFT analysis can be performed.
Fig. \ref{figure: FPFT 5DOF DO} shows the PDFs of FPFT for each DOF when the input force corresponds to 20FT.
It can be observed that the predicted PDFs are similar to the actual PDFs, thus demonstrating the applicability of the surrogate model for reliability analysis.
This case study is a perfect scenario where if the user wishes to obtain displacement at $N^{th}$-DOF, they can use the proposed surrogate model to obtain just that without producing the results for all the other DOFs, thus saving the computational time as well as expensive server space.
This is possible because displacement at each DOF is trained and mapped to input force individually.
The proposed surrogate model thus, in a sense, has the ability to decouple the $N$-DOF system without actually changing any of its dynamics or dropping any information.
Another advantage of individually training the surrogate model is that if the MSE at any particular DOF is high, initial samples or PPS at that particular DOF only can be changed, thus optimizing the computational requirements at the training stage.
\subsection{Case-III: $76$-DOF ASCE Benchmark Building}
\begin{figure}[ht!]
    \centering
    \includegraphics[width = \textwidth]{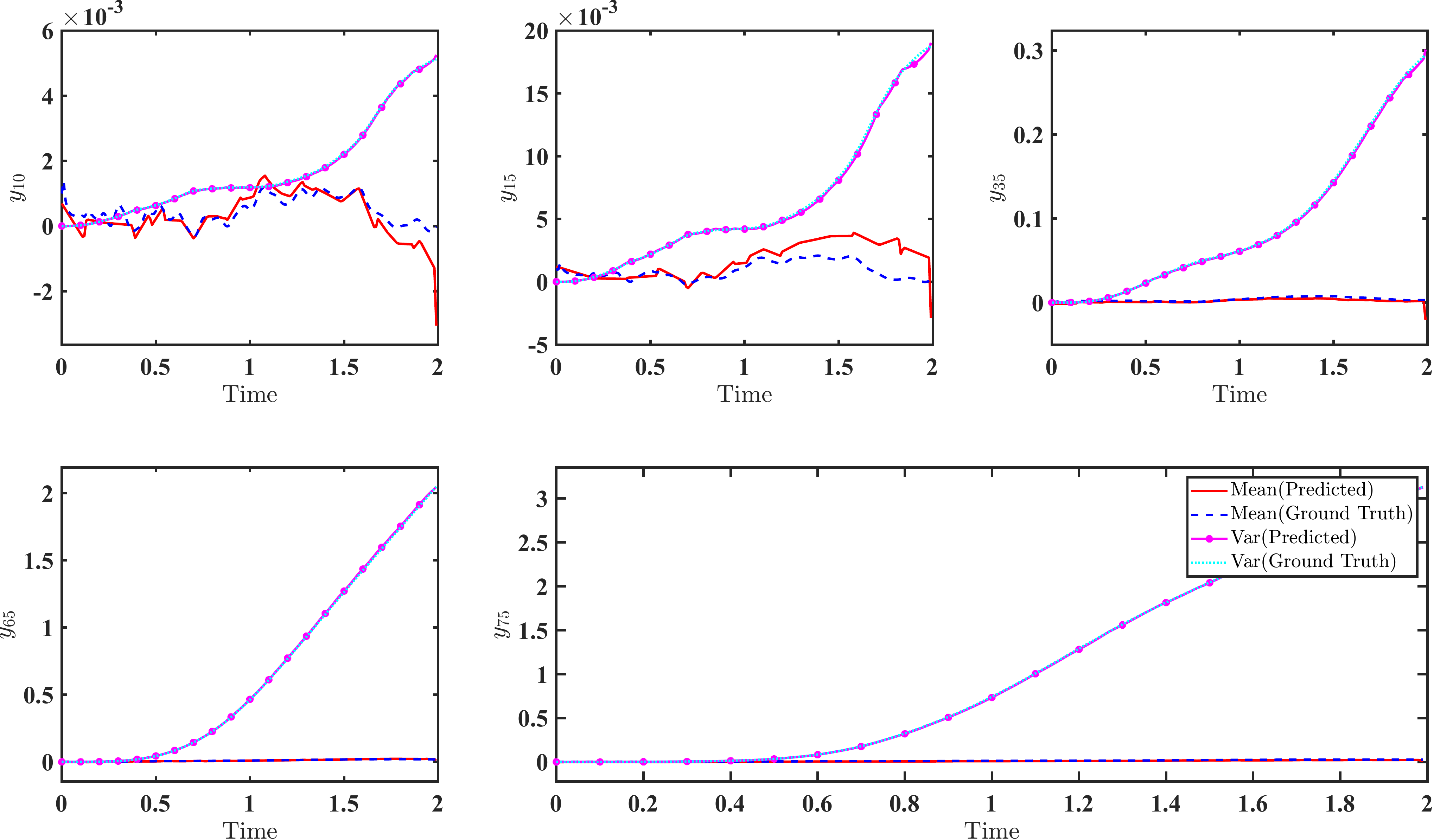}
    \caption{Predicted mean and variance of 10000 samples compared against the ground truth, for $76$-DOF linear ASCE benchmark (Case-III). Results are shown at $10^{th}$, $15^{th}$, $35^{th}$, $65^{th}$ and $75^{th}$ DOF. Training and testing has been done for input forces corresponding to 20 FT.}
    \label{figure: mean var plot 76DOF linear 20ft}
\end{figure}

\begin{figure}[ht!]
    \centering
    \includegraphics[width = \textwidth]{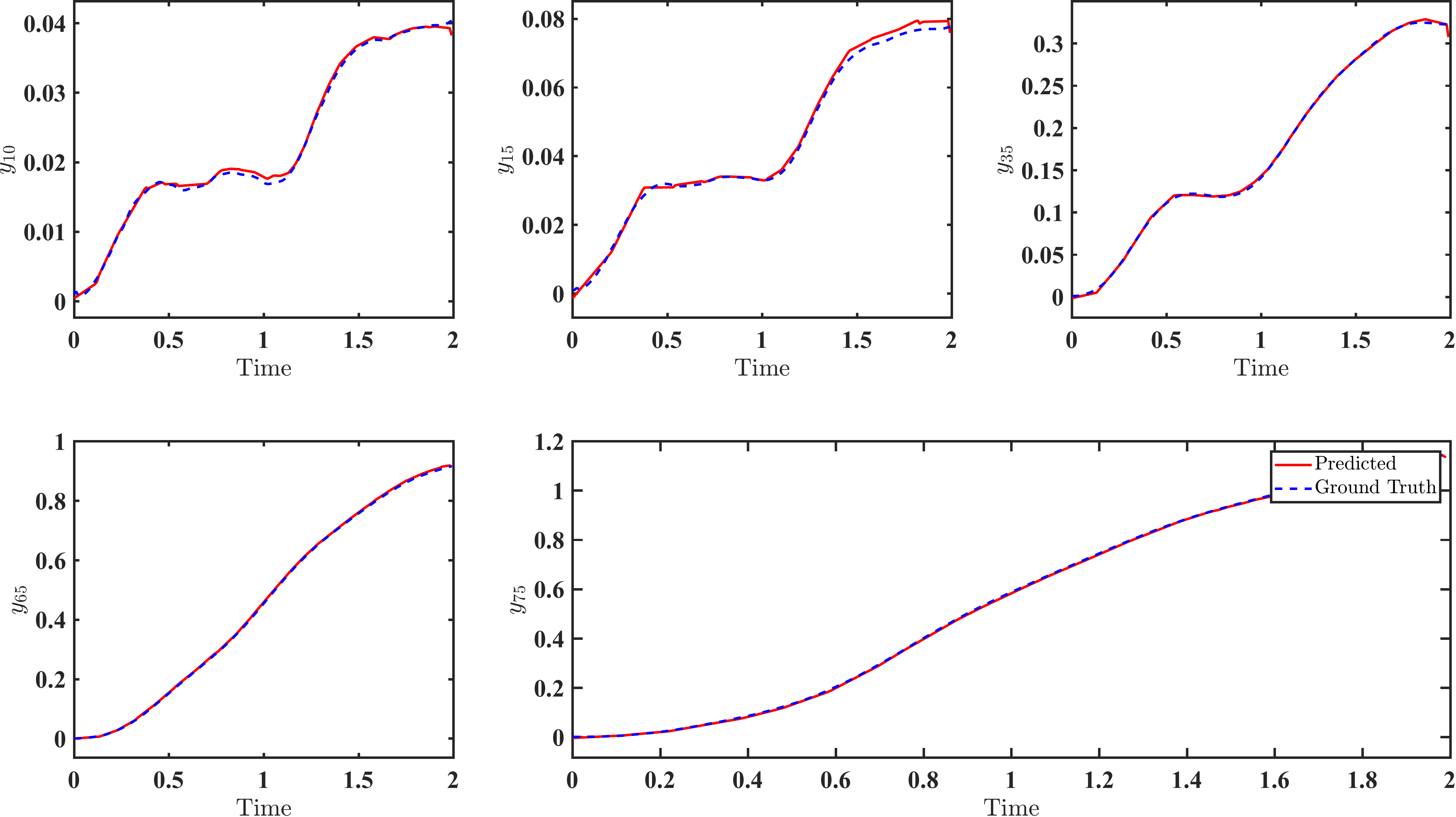}
    \caption{Predicted displacement for $1000^{th}$ sample compared against the ground truth, for $76$-DOF linear ASCE benchmark (Case-III). Results are shown at $10^{th}$, $15^{th}$, $35^{th}$, $65^{th}$ and $75^{th}$ DOF. Training and testing has been done for input force corresponding to 20 FT.}
    \label{figure: 1000th sample plot 76DOF linear 20ft}
\end{figure}

This case study aims to showcase the scalability of the proposed approach by applying the concept of surrogate model to a base excited $76$-DOF building, parameters for which are selected based on the ASCE benchmark \cite{yang2004benchmark,nayek2019gaussian}.
The differential equation for the system under consideration is as follows:
\begin{equation}
    \mathbf M \bm{\ddot X} + \mathbf C \bm{\dot X} + \mathbf K \bm X = -\mathbf M\bm I_f f,
    \label{equation: 76 dof linear system}
\end{equation}
where $I_f$ is the influence vector, and $f$ is the random ground acceleration, computed using Eq. \eqref{equation: force fourier}.
Initial displacements of 0.001 and initial velocities of 0.005 were also considered at all DOFs. 
Keeping DeepONet architecture same as that used in previous cases, initial samples for this case study are taken equal to 400 with 100 PPS each.
Similar to the previous MDOF case, $f$ will be mapped in parallel to displacements, and hence 76 different trained models will be available.
This brings with it the advantage of training displacements at only required DOFs, thus saving computational cost and time.
Fig. \ref{figure: mean var plot 76DOF linear 20ft} shows the plot comparing mean and variance of predicted displacements against the ground truth at $10^{th}$, $15^{th}$, $35^{th}$, $65^{th}$ and $75^{th}$ DOF when tested and trained for 20 FT and Fig. \ref{figure: 1000th sample plot 76DOF linear 20ft} shows the results for one particular sample ($1000^{th}$ sample).
Results show that the predicted values closely follow the ground truth, thus demonstrating the predictive capabilities of the proposed approach.

\begin{figure}[ht!]
    \centering
    \includegraphics[width = \textwidth]{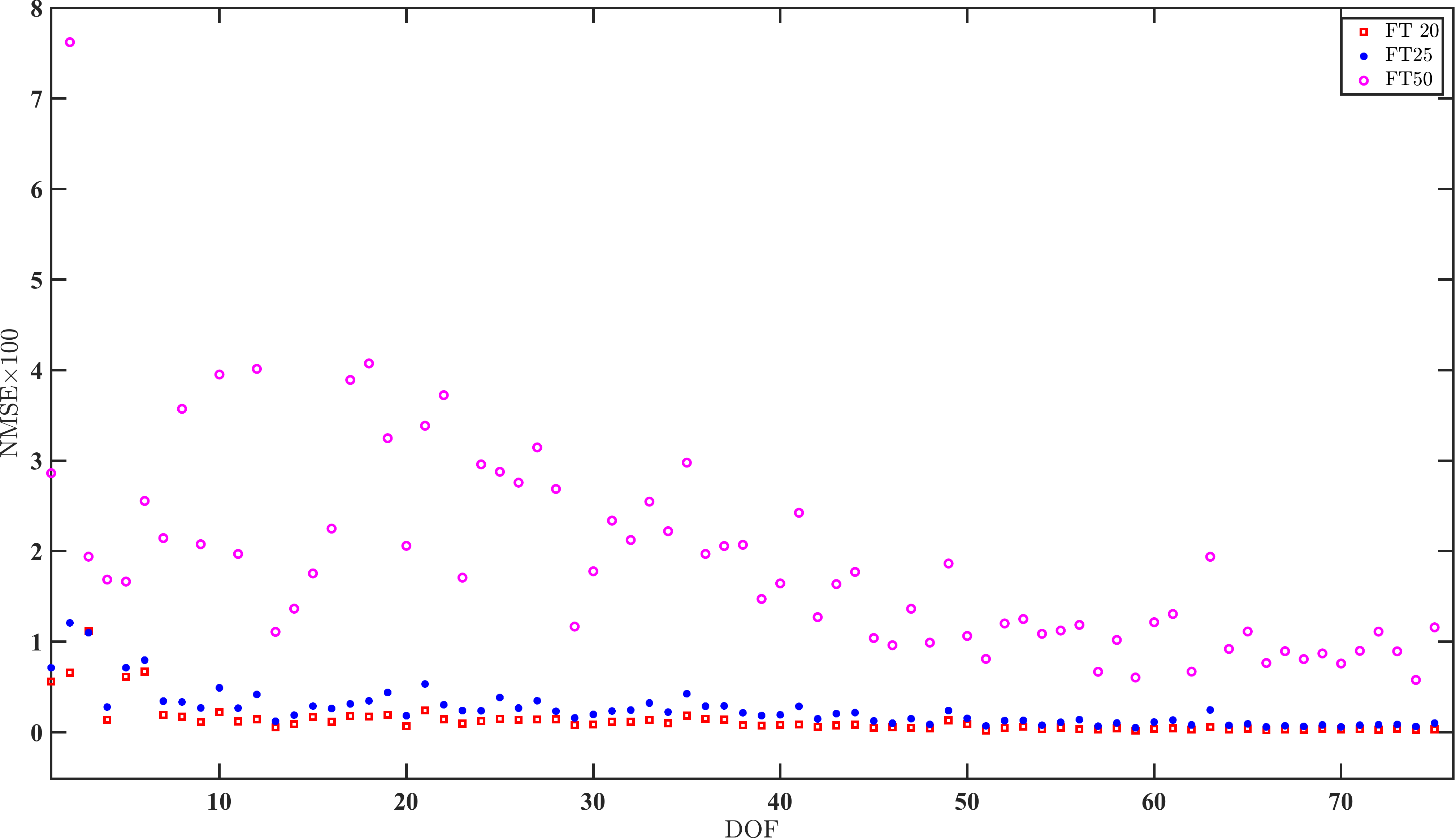}
    \caption{Percentage NMSE for 10000 samples. Model has been trained for input forces corresponding to 20 FT and testing has been done for 20 FT, 25 FT and 50 FT.}
    \label{figure: MSE 76DOF Linear}
\end{figure}

Fig. \ref{figure: MSE 76DOF Linear} shows the percentage normalized mean square error (NMSE) for case when surrogate model is trained for input forces corresponding to 20 FT and testing is carried out for 20 FT, 25 FT and 50 FT.
The results produced show the percentage NMSE error stays well within $5\%$ for the majority of cases and well within $2\%$ for 20 FT and 25 FT. NMSE has been calculated as follows:
\begin{equation}
    NMSE = \frac{1}{N_s}\mathlarger{\mathlarger{\sum}}\limits_{i = 1}^{N_s}\frac{||predicted-actual||^2_2}{||actual||^2_2},
\end{equation}
where $N_s$ is the number of samples.

\begin{figure}[ht!]
    \centering
    \includegraphics[width = \textwidth]{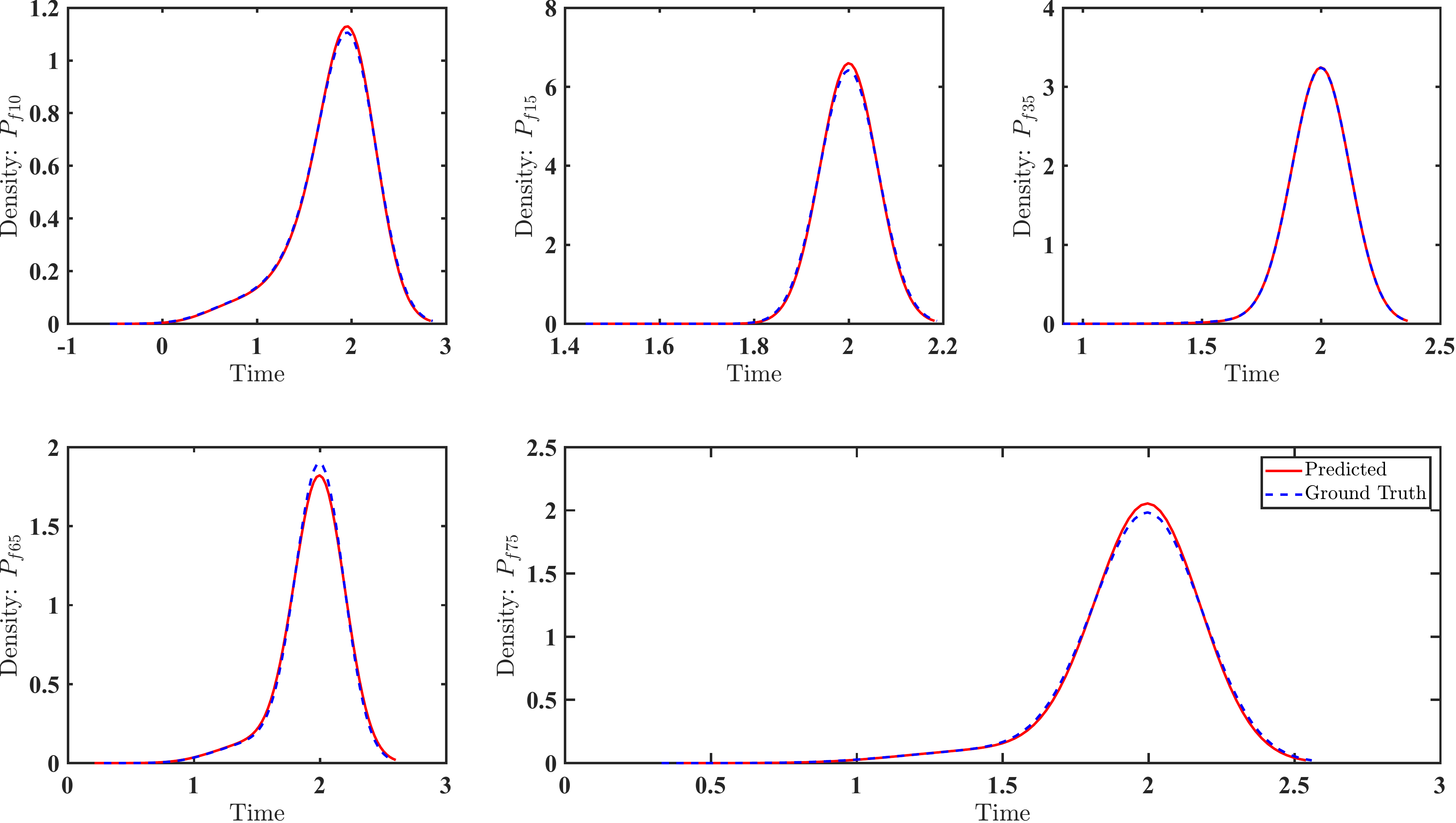}
    \caption{PDFs of FPFT obtained for Case-III at $10^{th}$, $15^{th}$, $35^{th}$, $65^{th}$ and $75^{th}$ DOF. Training and testing are done for input forces corresponding to 20FT.}
    \label{figure: FPFT 76DOF DO}
\end{figure}

Now, for reliability analysis, Fig. \ref{figure: FPFT 76DOF DO} shows the results for FPFT analysis at $10^{th}$, $15^{th}$, $35^{th}$, $65^{th}$ and $75^{th}$ DOF.
The results closely follow the ground truth for the $76$-DOF building, thus demonstrating the scalability of the proposed approach. Training and testing for the reliability analysis is done for input forces corresponding to 20FT.
\subsection{Case-IV: $76$-DOF nonlinear system}
To further showcase the robustness and explore the advantages of the proposed surrogate model, $76$-DOF building used in the previous case study is modified to include a Bouc Wen oscillator at first DOF.
It is assumed that the nonlinear terms will prominently affect the differential equation at first DOF only, and appropriate changes have been made in the differential equations of the system.
For data simulation using Runge-Kutta scheme, a sampling frequency of 12500$Hz$ was adopted, but data provided for training was sampled at a frequency of 100$Hz$.
The rest of the DeepONet architecture was kept similar to previous case studies, with initial samples here also being 400 with 100 PPS.

\begin{figure}[ht!]
    \centering
    \includegraphics[width = \textwidth]{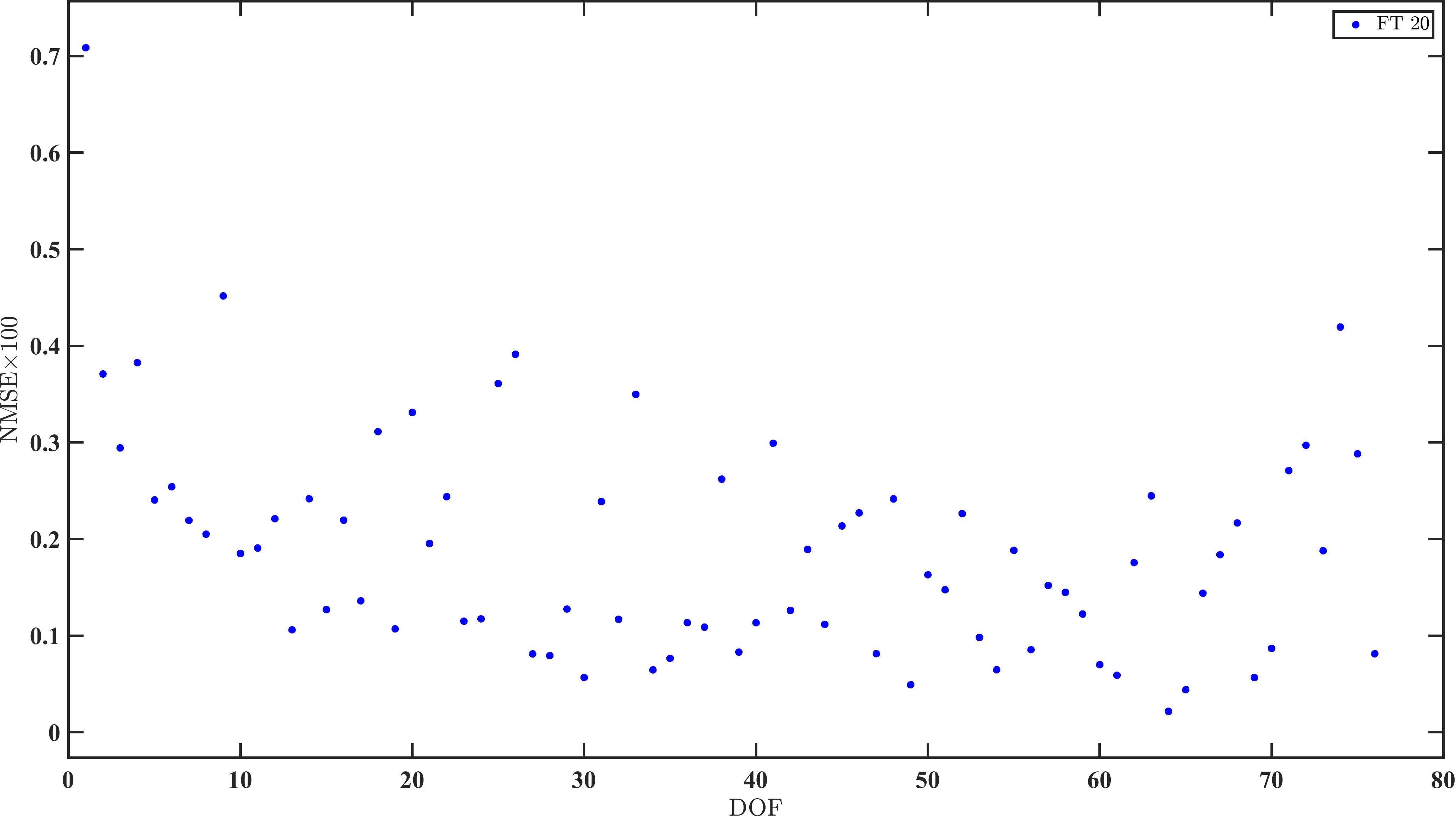}
    \caption{Percentage NMSE for 10000 samples (Case-IV). Model has been trained and tested for 20 FT.}
    \label{figure: NMSE 76DOF NL linear}
\end{figure}

\begin{figure}[ht!]
    \centering
    \includegraphics[width = \textwidth]{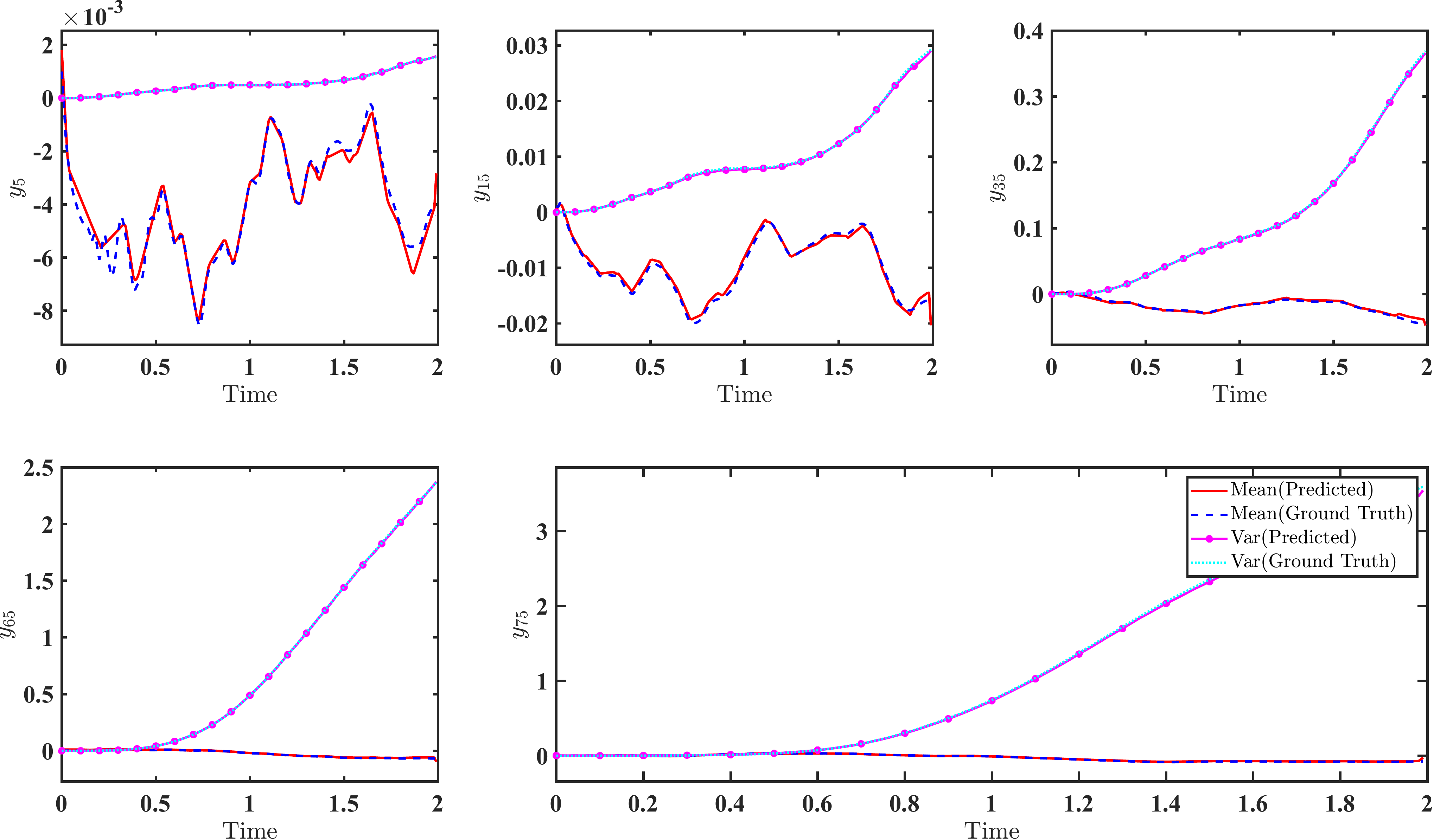}
    \caption{Predicted mean and variance of 10000 samples compared against the ground truth, for $76$-DOF nonlinear building (Case-IV). Results are shown at $5^{th}$, $15^{th}$, $35^{th}$, $65^{th}$ and $75^{th}$ DOF. Training and testing has been done for input forces corresponding to 20 FT.}
    \label{figure: 20FT 76DOF NL mean var}
\end{figure}

Fig. \ref{figure: NMSE 76DOF NL linear} shows the percentage NMSE for 10000 samples at each DOF. The results produced show that the NMSE is well below 1\% thus giving a reasonably accurate picture of the actual system.
To further showcase the performance of the surrogate model, Fig. \ref{figure: 20FT 76DOF NL mean var} compares the predicted mean and variance with the ground truth.
As expected, the model is able to predict with great accuracy, the displacements for the new input forces.

\begin{figure}[ht!]
    \centering
    \includegraphics[width = \textwidth]{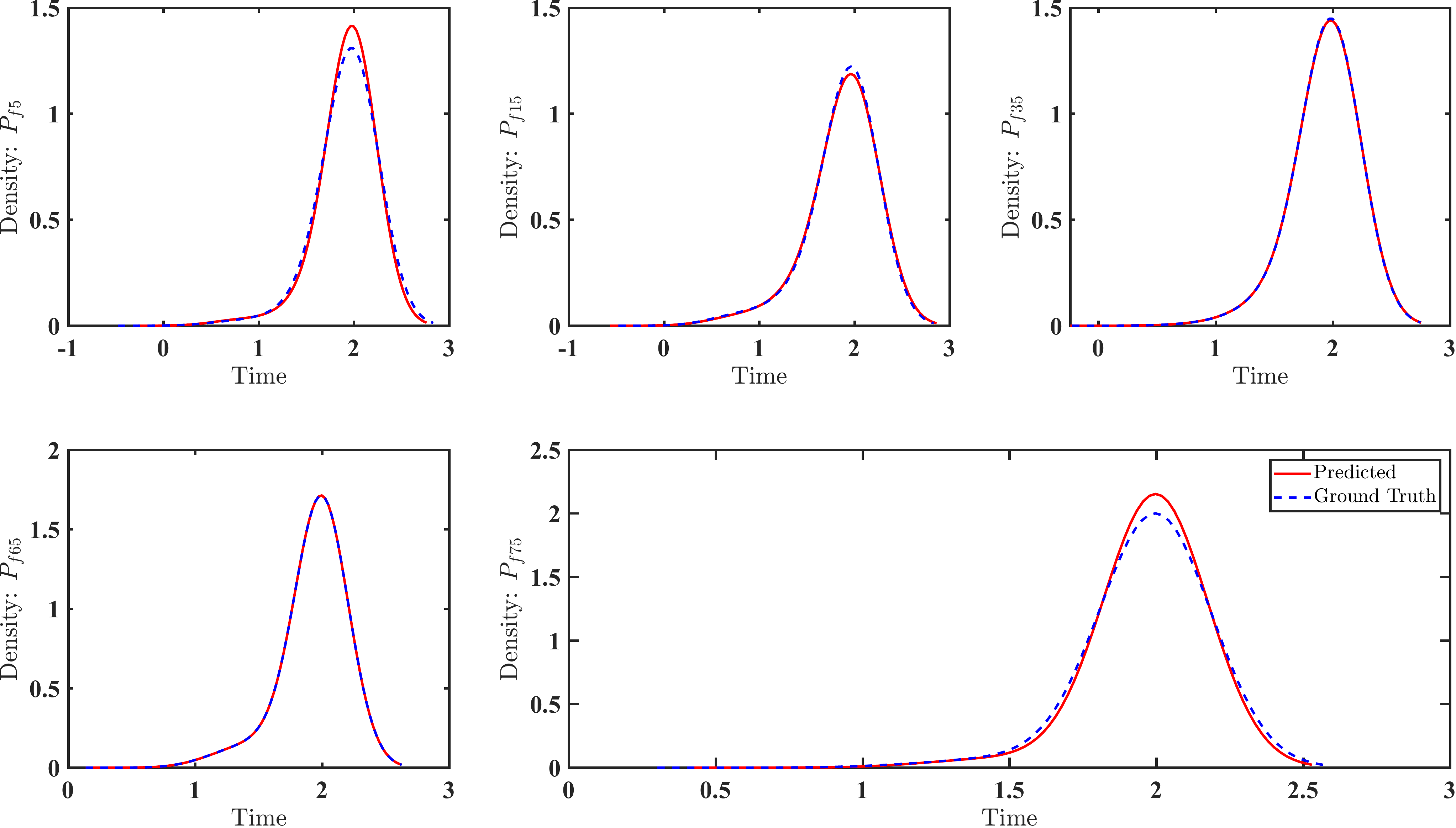}
    \caption{PDFs of FPFT obtained for Case-IV at $5^{th}$, $15^{th}$, $35^{th}$, $65^{th}$ and $75^{th}$ DOF. Training and testing are done for input forces corresponding to 20FT.}
    \label{figure: 20FT 76DOF NL RA}
\end{figure}

Similar to previous case studies, FPFT PDFs have been computed and are plotted in Fig. \ref{figure: 20FT 76DOF NL RA} for $5^{th}$, $15^{th}$, $35^{th}$, $65^{th}$ and $75^{th}$ DOF.
The predicted PDFs closely follow the ground truth, thus enabling the designer to make well-informed decisions.
Note, that the displacements produced for Case-III and Case-IV, $76$-DOF buildings have high magnitudes for top stories as no control structure have been included in the analysis and despite this the surrogate model is still able to predict the displacements with great accuracy.
The Python source code for the case studies discussed above are available at {\color{blue}link to be added here later}.
\section{Conclusion}\label{section: Conclusion}
Reliability analysis is an integral part of any structural design, especially for structures subjected to uncertain loading conditions, and to perform a robust reliability analysis, ample data representing the actual system is often needed.
To achieve this, help of computer aided simulations is taken, as it is possible to test a wide variety of cases with little modification to underlying code.
But generating high-fidelity data has associated with it immense computational cost, especially for complex nonlinear dynamical systems. In this paper, we propose a surrogate model based technique to perform reliability analysis using neural networks, specifically DeepONet architecture.
Unlike conventional machine learning models, DeepONet learns a function to function mapping and hence, is ideally suited for time-dependent reliability analysis and uncertainty quantification problems subjected to loading uncertainty. 
The proposed framework enables efficient reliability analysis and uncertainty quantification subjected,  and also provides ZSL capabilities.
Using the proposed framework, it is possible to predict only the required data at coarser time steps, thus saving both expensive server space and computational cost.
Authors note that time will be spent on training the surrogate model, but the time saved in the prediction stage will sufficiently compensate for the time spent in training.
The proposed framework will especially benefit applications where the data simulation takes enormous amounts of time.

As a proof of concept, several case studies have been discussed above, testing the surrogate model against a variety of dynamical systems and assessing the use of DeepONet architecture for reliability analysis of nonlinear dynamical systems.
Case-I and II demonstrate the surrogate model's capability to model nonlinear oscillators, while Case III and IV showcase the scalability of the proposed framework.
Although at this stage, the predictions have been made for $10^4$ samples, in a practical scenario, $10^5$ or even more test samples may be required, and in those cases, the proposed algorithm can be immensely helpful in saving simulation time.
The results produced in the case studies represent the actual system with good accuracy.
Initial samples were kept limited so as to showcase the performance of surrogate model with less training data, and from the case studies, it can be observed that good results are obtained.

Based on the requirements of the system under consideration, DeepONet architecture can be easily tweaked, and appropriate results can be obtained. 
Some key points to note include the following. PPS selected should be uniformly distributed in the domain of the function for best training results.
Number of nodes and hidden layers may be increased if the length of individual samples in the training data are large.
As a rule of thumb, we recommend increasing the number of hidden layers first, before moving on to nodes per layer as this approach mitigates some chances of over-fitting.
Apart from this, importance of normalization in any NN architecture can not be overstated, and hence the data being given as the training input should be appropriately normalized.
Although the choice of normalization will depend on the application and type of data, we propose using Keras \textit{preprocessing normalization layer} \cite{Normalization}.
Min-max scaling function was also tested, and satisfactory results were obtained.
Overall, it is safe to conclude that DeepONet does a decent job in time-dependent reliability analysis and uncertainty quantification subjected to stochastic loading.
\section*{Acknowledgment}
SG acknowledges the support received from the Ministry of Education in the form of Ph.D. scholarship.
SC acknowledges the financial support received from IIT Delhi.

\end{document}